\documentclass[conference]{IEEEtran}
\IEEEoverridecommandlockouts
\usepackage{amsmath,amsthm,amssymb,amsfonts}
\usepackage{mathtools}
\usepackage{bibentry}
\nobibliography*
\usepackage[dvipsnames]{xcolor}
\usepackage{bm}
\usepackage{endnotes}
\usepackage{pdflscape}

\usepackage{amssymb,amsfonts}
\usepackage{caption}
\usepackage{subcaption}
\usepackage{graphicx}

\usepackage{url}
\usepackage{enumerate}
\usepackage{multicol}
\usepackage{multirow}
\usepackage{mathtools}
\usepackage{tabularray}
\usepackage{cals}
\usepackage{algorithm}
\usepackage{algpseudocode}
\usepackage{color}
\usepackage{array}
\usepackage{multirow}
\usepackage{eufrak}
\usepackage{comment}
\usepackage{tikz}
\usetikzlibrary{arrows.meta, positioning}
\usetikzlibrary{shapes}
\usetikzlibrary{calc}

\definecolor{Blue}{HTML}{000000}

\newcommand{\mathbbm}[1]{\text{\usefont{U}{bbm}{m}{n}#1}}

\usepackage{amsthm}
\theoremstyle{plain}
\newtheorem{theorem}{Theorem}
\newtheorem{lemma}{Lemma}

\newtheorem{proposition}{Proposition}

\definecolor{hilo_yellow}{RGB}{255,154,0}

\usepackage{cite}
\usepackage{amsmath,amssymb,amsfonts}
\usepackage{graphicx}
\usepackage{textcomp}
\usepackage{xcolor}
\def\BibTeX{{\rm B\kern-.05em{\sc i\kern-.025em b}\kern-.08em
    T\kern-.1667em\lower.7ex\hbox{E}\kern-.125emX}}

\usepackage{float}
\usepackage[dvipsnames]{xcolor}
\usepackage{tikz}

\usepackage{array} 
\usepackage{booktabs}  
\usepackage{multirow}
\usepackage{colortbl}

\usepackage{siunitx} 

\usepackage[hidelinks]{hyperref}

\begin{document}

\title{
Sliced-Wasserstein Distance-based Data Selection
\thanks{This work was funded by Mitacs, the Natural Sciences and Engineering Research Council of Canada (NSERC), and Services Hilo under Grants RGPIN-2023-04235 and IT35303 \& IT38517. Corresponding authors: \{\texttt{julien.pallage}, \texttt{antoine.lesage-landry}\}@\texttt{polymtl.ca}.}
}

\author{
    \IEEEauthorblockN{Julien Pallage}
    \IEEEauthorblockA{        
        Department of Electrical Engineering\\
        Polytechnique Montréal,
        Mila \& GERAD}
    \and
    \IEEEauthorblockN{Antoine Lesage-Landry}
    \IEEEauthorblockA{
        Department of Electrical Engineering\\
        Polytechnique Montréal,
        Mila \& GERAD}
}

\maketitle

\thispagestyle{plain} 
\pagestyle{plain}


\begin{abstract}
We propose a new unsupervised anomaly detection method based on the sliced-Wasserstein distance for training data selection in machine learning approaches. Our filtering technique is interesting for decision-making pipelines deploying machine learning models in critical sectors, e.g., power systems, as it offers a conservative data selection and an optimal transport interpretation. To ensure the scalability of our method, we provide two efficient approximations. The first approximation processes reduced-cardinality representations of the datasets concurrently. The second makes use of a computationally light Euclidian distance approximation.  Additionally, we open the first dataset showcasing localized critical peak rebate demand response in a northern climate. We present the filtering patterns of our method on synthetic datasets and numerically benchmark our method for training data selection. Finally, we employ our method as part of a first forecasting benchmark for our open-source dataset.
\end{abstract}



\section{Introduction}\label{sec:intro}

Virtual power plants (VPPs)~\cite{pudjianto2007virtual} are networks of distributed energy resources (DERs), e.g., photovoltaic panels, electric vehicle batteries, thermostatically controlled loads, and other controllable grid-edge energy resources that are collectively managed and operated as a unified entity through a central control system. Instead of generating power from a single location like conventional power plants, VPPs mobilize and incentivize geographically dispersed generation, storage, and consumption to offer a vast range of grid services. 
VPPs rely on the forecast, the optimization, and the control of DERs to exploit their flexibility and meet grid needs with limited infrastructure change as well as minimal investments~\cite{zhang2019comprehensive}. The concept of VPP stands at the centrepiece of the grid decarbonization paradigm as it allows the integration of more renewable energy sources by helping mitigate their inherent intermittency with added flexibility.

Demand response (DR) is one of the many tools VPP operators utilize to gain flexibility. DR encourages customers power consumption pattern shifts by leveraging financial incentives~\cite{albadi2007DR}. It is enforced directly through load control or indirectly by promoting specific consumption patterns. DR programs include real-time pricing~\cite{Lijesen_2007_realtime_elasticity}, time-of-use pricing~\cite{Yang2013_ToU}, and critical peak rebates (CPRs)~\cite{6254817}. To be effective, DR programs can leverage historical data and machine learning~(ML) to fine-tune and optimize their services for current and future needs. Unfortunately, there is a general trade-off. More precise ML models in VPP applications should lead to better decision-making and facilitate renewable integration. Yet, more precise ML models often come with added complexity, which makes them riskier for deployment as interpretability, stability, and predictability may deteriorate. Finally, implementing a complete safe ML pipeline can sometimes be impossible for some large-scale VPP applications where computational resources are limited.

Dataset quality can highly influence the performance of ML models. Similarly to other internet of things (IoT) applications~\cite{liu2020data}, VPPs may suffer from heavy data corruption that arises from their reliance on sensors and telecommunication systems. The adversarial properties stemming from the combination of numerical errors, noise in sensor readings, telemetry issues, meter outages, and unusual extreme events can disrupt the prediction quality of ML models. Outlier filtering is thus primordial in a reliable ML pipeline, especially when limited regularization mechanisms are implemented during training. In this work, we tackle the challenges of training ML models with unfiltered data and propose a new unsupervised outlier filter leveraging the sliced-Wasserstein (SW) distance~\cite{bonnotte2013unidimensional}. We illustrate the performance of this filter for data selection.

We also address the lack of open-source datasets showcasing DR mechanisms in northern climates, namely CPR, by releasing two years of aggregated consumption data for customers participating in a localized CPR (LCPR) program in Montréal, Québec, Canada\footnote{Available at \url{https://github.com/jupall/lcpr-data}, and \url{https://donnees.hydroquebec.com/explore/dataset/consommation-clients-evenements-pointe}}. These customers are spread in three adjacent distribution substations. With it, we hope to stimulate research on trustworthy ML~\cite{chatzivasileiadis2022machine} models for DR applications. We leverage our outlier filter method to present a first benchmark for the prediction of the localized energy consumption of LCPR participants on our open-source dataset.

We now review the literature associated to unsupervised outlier filtering for ML and provide some context on CPRs. 

\emph{Outlier filtering.}
Unsupervised outlier filtering or anomaly detection (AD) methods are preferable as they do not need human-made labels and their hyperparameters can be tuned simultaneously with other ML models included in the loop. The main unsupervised AD methods~\cite{goldstein2016AD} include local outlier factor (LOF)~\cite{Breuning2000LOF}, isolation forest~\cite{Liu2008IsolationForest}, $k$-nearest neighbours (KNN)~\cite{angiulli2002knn}, connectivity-based outlier factor~\cite{Tang2002CLOF}, and one-class support vector machine (SVM)~\cite{Amer2013OneClassSVM}. These methods either use clustering or local density to assign an outlier score. 

We are interested in optimal transport-based (OT) metrics for AD because of the intuition it adds to the unsupervised data selection phase. Reference~\cite{ducoffe2019anomaly} proposed a Wasserstein generative adversarial network for AD tasks and~\cite{wang2024WOOD} designed a differentiable training loss function using the Wasserstein metric for deep classifiers. To the best of our knowledge, no unsupervised OT-based method has been proposed yet for AD.

\emph{Localized critical peak rebates.} 
%
Residential customers enrolled in a CPR program receive financial compensation during pre-specified time periods, referred to as \emph{challenges} in Québec's power system setting, for reducing their energy consumption with respect to their expected baseload~\cite{mercado2014enabling}. CPR programs are purely voluntary and virtually penalty-free. 
We consider a variation of CPR, viz., localized CPR (LCPR), in which the events are called for localized relief in the grid instead of being cast for the whole system. LCPR can diversify the types of services offered by typical CPR programs, e.g., they can alleviate stress on local equipment like substation transformers~\cite{feng2024gridedge} or they can be paired with generation forecasts of DERs to balance demand and generation during peak hours. LCPR is underexplored in the literature and is a valuable application to benchmark trustworthy ML models. Indeed, the higher spatial granularity, the critical aspect of the task, the dependence on behavioural tendencies, and the lower margin for error require the deployment of forecasting models that offer performance guarantees \cite{Venzke2021Worstcase}, robustness to noise~\cite{Chen2020_DRO}, physical constraint satisfaction~\cite{Misyris2020PINN_PES}, a sense of prediction confidence~\cite{Jospine2022_handson_BNN}, or a combination of them~\cite{pallage2024wasserstein}. Being able to predict and utilize localized peak-shaving potential in the electrical grid, through programs similar to LCPR, could accelerate the integration of DERs.
To our knowledge, no published open-source datasets exhibit LCPR or CPR schemes in a northern climate. In sum, our contributions are as follows:
\begin{itemize}
    \item We propose a method, SW-based anomaly detection (\texttt{SWAD}), employing the SW distance to empirically assess the transportation cost of individual data samples with respect to the rest of the dataset. We motivate this method for training data selection and outlier filtering.
    \item We propose two scalable variations of \texttt{SWAD} that aim at a deployment on large datasets. The first method, smart split-\texttt{SWAD} (\texttt{sSWAD}), parallelizes computations by working with multiple reduced-cardinality representations (splits) of the original dataset. The second method, fast Euclidian  approximation AD (\texttt{FEAD}), uses an approximation to accelerate computations significantly. We motivate \texttt{FEAD} by proving that it is exactly equivalent to using the order-$1$ Wasserstein distance, and an upper bound on the order-$t$ Wasserstein distance. 
    \item We show the strengths and weaknesses of our methods in several experiments. We start by qualitatively exposing the filtering paradigm of our methods. We then numerically compare our methods to classical AD algorithms for training data selection.
    \item We open the first dataset that demonstrates LCPR mechanisms in a northern climate and use our filtering method in the making of a first forecasting benchmark. 
\end{itemize}

We remark that an unpublished preliminary version of this work has been presented at NeurIPS's CCAI Workshop~\cite{pallage2024sliced}. The rest of the paper has the following structure. In Section~\ref{sec:data_pipeline}, we introduce an example of a simplified data pipeline in IoT applications like VPPs and elaborate on the different corruption sources. In Section~\ref{sec:swfilter_main}, we present our new empirical methods for data filtering. In Sections~\ref{sec:swfilter_qualitative} and~\ref{sec:swfilter_experiments}, we present a qualitative study and the different numerical experiments, respectively. Finally, we present our open-source dataset in Section~\ref{sec:open_source} with closing remarks in Section~\ref{sec:swfilter_closing}.

\section{Data corruption in data-intensive applications}\label{sec:data_pipeline}

Data corruption can happen at different stages of the data pipeline. To support our discussion, we present a stylized version of this pipeline in Figure~\ref{fig:data_pipeline} for an application involving communicating sensors and controllers, e.g., VPPs~\cite{ruan2024data}. We refer readers to the work of \cite{reis2022fundamentals} and \cite{kleppmann2017designing} for a more global and in-depth dive into the concepts of data engineering and the requirements of reliable data-intensive applications. 

\tikzstyle{double_arrow} = [dotted,{Stealth[scale=1.2]}-{Stealth[scale=1.2]}]

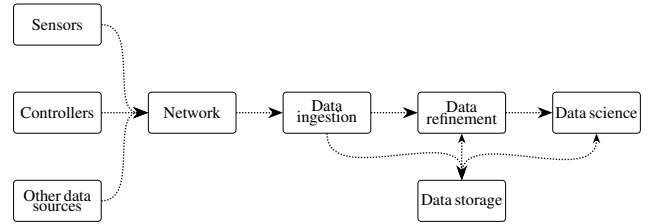
\begin{figure}[H]
    \centering

\resizebox{0.95\columnwidth}{!}{
\begin{tikzpicture}[
    node distance=2cm,
    every node/.style={rounded corners, align=center, text width=10em,, minimum size=5em,anchor=center},
    decision/.style={diamond, draw, text width=3em, align=center},
    arrow/.style={-{Stealth[scale=1.8]}, dotted,ultra thick}
]
\node[draw,] (sensors) {\huge Sensors};
\node[draw,below=2cm of sensors] (controllers) {\huge Controllers};
\node[draw,below=2cm of controllers] (extdata) {\huge Other data sources};
\node[draw, right= 2cm of controllers] (net) {\huge Network};
\node[draw,right=2cm of net] (acquisition) {\huge Data ingestion};
\node[draw,right=2cm of acquisition] (processing) {\huge Data refinement};
\node[draw,right=2cm of processing] (exploitation) {\huge Data science};
\node[draw,below=2cm of processing] (storage) {\huge Data storage};

\draw[arrow] (sensors.east)  to [out=0,in=180] (net.west) ;
\draw[arrow] (controllers.east)  to [out=0,in=180] (net.west) ;
\draw[arrow] (extdata.east)  to [out=0,in=180] (net.west) ;

\draw[arrow] (net.east)  to [out=0,in=180] (acquisition.west);
\draw[arrow] (acquisition.east)  to [out=0,in=180] (processing.west);
\draw[arrow] (processing.east)  to [out=0,in=180] (exploitation.west);
\draw[arrow] (acquisition.south)  to [out=-90,in=90] (storage.north);
\draw[double_arrow,ultra thick] (processing.south)  to [out=-90,in=90] (storage.north);
\draw[double_arrow,ultra thick] (exploitation.south)  to [out=-90,in=90] (storage.north);

%


\end{tikzpicture}
}
    \caption{Stylized data pipeline}
    \label{fig:data_pipeline}
\end{figure}

We now provide a general layout of the procedure, starting with data ingestion. Data ingestion is the process of collecting and importing raw data from various sources into a centralized system. This data can come from IoT devices, sensors, communicating controllers, web resources, or external databases and often requires pre-processing to remove duplicates, handle missing values, and standardize formats. Once ingested, data refinement ensures that the raw data is cleaned, structured, and transformed to be usable. This step includes filling in missing values with extrapolation methods, normalizing formats, correcting errors, and filtering out anomalies. After refinement, data science extracts knowledge from the processed data using statistical analysis, ML, and domain-specific techniques. This allows for pattern detection, predictive modelling, and data-driven decision-making~\cite{reis2022fundamentals}. 

Corruption can occur throughout the pipeline. At the edge, sensors may produce noisy readings due to environmental interference, poor calibration, or hardware failures. As data moves through the network, telemetry errors, packet loss, and connectivity issues like Internet or power outages can create gaps in the data. During ingestion, duplicates and timestamp misalignments can also introduce inconsistencies. Refinement itself can lead to numerical errors, incorrect transformations, or loss of precision. Finally, other sources like outlier events, computational bugs, human design errors, and cascading failures can mislead the analysis and introduce further biases~\cite{kleppmann2017designing}.

Even though different data validation mechanisms can be implemented at each stage to limit data corruption, it is important to integrate data selection methods into the ML process to promote a foolproof procedure and enhance reliability.

\section{Sliced-Wasserstein-based anomaly detection}\label{sec:swfilter_main}

We now devise our SW-based methods for data selection.

\subsection{Preliminaries}
The Wasserstein distance is a metric that provides a sense of distance between two distributions. Also called earth mover's distance, it can be conceived as the minimal effort it would take to displace a pile of a weighted resource to shape a specific second pile~\cite{panaretos2019statistical}. Let $\mathcal{Z} \subseteq \mathbb{R}^d$, $d \in \mathbb{N}$, be the uncertainty set of the data, and let $\mathcal{P}(\mathcal{Z})$ be the set of all distributions supported on $\mathcal{Z}$~\cite{Chen2020_DRO}. The order-$t$ Wasserstein distance with respect to some norm $\|\cdot\|$ between distributions $\mathbb{U} \in \mathcal{P}(\mathcal{Z})$ and $\mathbb{V} \in \mathcal{P}(\mathcal{Z})$ is defined as: 
\[W_{\|\cdot\|,t}(\mathbb{U}, \mathbb{V}) = \left( \inf_{\pi \in \mathcal{J}(\mathbb{U}, \mathbb{V})} \int_{\mathcal{Z}\times\mathcal{Z}} \|\mathbf{z}_1 - \mathbf{z}_2\|^t \mathrm{d}\pi(\mathbf{z}_1,\mathbf{z}_2)\right)^{\frac{1}{t}},
\]
where $\mathcal{J}(\mathbb{U}, \mathbb{V})$ represents all possible joint distributions $\pi$ between $\mathbb{U}$ and $\mathbb{V}$, and $\mathbf{z}_1$ and $\mathbf{z}_2$ are the marginals of $\mathbb{U}$ and $\mathbb{V}$, respectively. 
In general, computing the Wasserstein distance is of high computational complexity~\cite{kolouri2019generalized} except for some special cases. For example, the Wasserstein distance for one-dimensional distributions has a closed-form solution that can be efficiently approximated \cite{bonnotte2013unidimensional}.

The SW distance is a metric that makes use of this property by calculating infinitely many linear projections of the high-dimensional distribution onto one-dimensional distributions and then computing the average of the Wasserstein distance between these one-dimensional representations~\cite{bonnotte2013unidimensional}. Interestingly, it possesses similar theoretical properties to the original Wasserstein distance while being computationally tractable~\cite{bonnotte2013unidimensional}. The order-$t$ SW distance with respect to some norm $\|\cdot\|$ can be approximated using a Monte Carlo method employing direction samples $\bm{\theta}_l$, $l=1,2,\ldots, L$ uniformly distributed on the $d$-dimensional unit sphere $\mathcal{S}^{d-1}_1$. The finite approximation is defined as: 
\begin{equation*}S W_{\|\cdot\|,t}\left(\mathbb{U}, \mathbb{V} \right) \approx\left(\frac{1}{L} \sum_{l=1}^L W_{\|\cdot\|,t}\left(\mathfrak{R} \mathbb{U}\left(\cdot, \bm{\theta}_l\right), \mathfrak{R} \mathbb{V}\left(\cdot, \bm{\theta}_l\right)\right)^t\right)^{\frac{1}{t}} \hspace{-0.5em},
\end{equation*} 
where $\mathfrak{R} \mathbb{U}(\cdot, \bm{\theta}_l)$ and $\mathfrak{R} \mathbb{V}(\cdot, \bm{\theta}_l)$ are the Radon transform of distribution functions $\mathbb{U}$ and $\mathbb{V}$, respectively, i.e., linear projections onto the direction 
$\bm{\theta}_l \in \mathcal{S}^{d-1}_1 = \{\bm{\theta} \in \mathbb{R}^d \, | \, \|\bm{\theta}_l\|_2 = 1\}$ for $l \in [\![L]\!]$~\cite{kolouri2019generalized}.




The SW distance has some interesting relation with the Wasserstein distance as demonstrated by~\cite{bonnotte2013unidimensional}.
\begin{theorem}[Equivalence of ${SW}_{\|\cdot\|_t,t}$ and ${W}_{\|\cdot\|_t,t}$~\cite{bonnotte2013unidimensional} ]\label{theo:equivSW}
    There exists a constant $0<C({d, t})<+\infty$ such that, for all $\mathbb{U}, \mathbb{V} \in \mathcal{P}(\mathcal{B}(0, R))$, where $\mathcal{B}(0, R)$ is the closed ball of radius $R>0$ in $\mathbb{R}^d$ centred at the origin:
\begin{align*}
    {SW}_{\|\cdot\|_t,t}(\mathbb{U}, \mathbb{V})^t &\leq c({d, t}) {W}_{\|\cdot\|_t,t}(\mathbb{U}, \mathbb{V})^t \\
    &\leq C({d, t}) R^{t-1 /(d+1)} {SW}_{\|\cdot\|_t,t}(\mathbb{U}, \mathbb{V})^{1 /(d+1)},
\end{align*}
where $c({d, t})=\frac{1}{d} \int_{\mathcal{S}^{d-1}_1}\|\theta\|_t^t \mathrm{~d} \theta \leq 1$.
\end{theorem} 
Theorem~\ref{theo:equivSW} effectively shows that the SW metric can be used to bound the Wasserstein distance to the power of $t$.

\subsection{Sliced-Wasserstein data selection}
We now utilize the finite approximation of the SW distance to formulate a simple outlier filter. Let $\mathcal{D} = \bigcup_{i=1}^N\{\mathbf{z}_i\}$ be a dataset of $N$ independent samples $\mathbf{z}_i \in \mathbb{R}^d$. Consider an empirical distribution $\hat{\mathbb{P}}_N = \frac{1}{N} \sum_{i=1}^N \delta_{\mathbf{z}_i}(\mathbf{z})$, where $\delta_{\mathbf{z}_i}(\cdot)$ is a Dirac delta function assigning a probability mass of one at each known sample $\mathbf{z}_i$. Let $\hat{\mathbb{P}}_{N-1}^{-\mathbf{z}_i}$ denote a variation of $\hat{\mathbb{P}}_N$ in which we remove sample $\mathbf{z}_i$. We propose an outlier filter by using a voting system in which we compare the SW distance between $\hat{\mathbb{P}}_N$ minus the outlier candidate and $\hat{\mathbb{P}}_N$ minus a random sample. Let $\mathcal{O} \subseteq \mathcal{D}$ and $\mathcal{I} \subseteq \mathcal{D}$ denote the sets of outlier and inlier samples, respectively, under the perspective of the filter; where $\mathcal{D} = \mathcal{O} \cup \mathcal{I}$. A vote is positive if the distance between the two reduced empirical distributions exceeds a threshold $\epsilon > 0$. A sample is labelled an outlier if the proportion of positive votes is greater or equal to a set threshold $p>0$. 
For compactness, we define $\Psi_{\|\cdot\|, t}(i,j,\epsilon) = \mathbbm{1}\left(SW_{\|\cdot\|,t} (\hat{\mathbb{P}}_{N-1}^{-\mathbf{z}_i} , \ \hat{\mathbb{P}}_{N-1}^{- \mathbf{z}_j})\geq \epsilon \right)$, where $\mathbbm{1}(\cdot)$ is the indicator function. The outlier set is then:
\begin{equation}\label{eq:swod}\tag{\texttt{SWAD}}
    \mathcal{O} = \left\{ \mathbf{z}_i \in \mathcal{D} \ \middle\vert \ p \leq \frac{1}{n} \hspace{-0.5em}\sum_{\substack{\mathbf{z}_j \sim \hat{\mathbb{P}}_{N-1}^{-\mathbf{z}_i}:\\ j\in [\![  n ]\!]}} \hspace{-0.5em} \Psi_{\|\cdot\|, t}(i,j,\epsilon),  \forall i \in [\![  N ]\!] \right\},
\end{equation}
where $n$ is the number of points used for the vote, $p \in [0,1]$ is the voting threshold required to label a sample as an outlier, and $[\![  N ]\!] \equiv \{1, 2, \ldots, N\}$. 

This filter is interesting for several reasons. It is unsupervised, purely analytical, and uses a well-known explainable distance function to filter data points that appear out-of-sample under the SW metric. The intuition is that we remove samples that are costly in the transportation plan when compared to other random samples of the same distribution. We can use the voting percentage to measure the algorithm's confidence in its labelling. It is also parallelizable, as seen in our implementation provided on our GitHub page\footnote{\url{https://github.com/jupall/swfilter}}. 

Unfortunately, this method does not scale well with large datasets: the SW distance computational burden increases when larger datasets are processed. We thus propose two methods to accelerate computations for real-world applications. 
\subsection{Smart splitting}
We start by proposing a \textit{smart} splitting variation. Instead of considering the whole dataset, we first use a clustering method, e.g., KNN, to obtain $K \in \mathbb{N}^*$ clusters. We then randomly split each cluster into $S\in \mathbb{N}^*$ splits and assemble $S$ reduced-cardinality representations of $\mathcal{D}$ by collecting one split from each cluster. We denote each split of the original dataset by $\mathcal{D}_s$ such that $\bigcup_{s=1}^S \mathcal{D}_s= \mathcal{D}$. This \textit{smart} splitting allows us to parallelize the outlier detection by treating each split concurrently. It is a viable option when multiple computation threads are available.

Abusing of notation, consider splits of the empirical distribution $\hat{\mathbb{P}}_{|\mathcal{D}_s|} = \frac{1}{|\mathcal{D}_s|} \sum_{\mathbf{z}_i \in \mathcal{D}_s} \delta_{\mathbf{z_i}}(\mathbf{z})$ where $|\mathcal{D}_s|$ denotes the cardinality of the split $s$, $\forall s \in [\![  S ]\!]$. For compactness, denote $\Psi_{\|\cdot\|, t,s}(i,j,\epsilon) = \mathbbm{1}\left(SW_{\|\cdot\|,t} (\hat{\mathbb{P}}_{|\mathcal{D}_s|-1}^{-\mathbf{z}_i},\hat{\mathbb{P}}_{|\mathcal{D}_s| -1}^{-\mathbf{z}_j} )\geq \epsilon \right)$. The methodology is now as follows: 
\begin{equation*}
\mathcal{O}_s = \left\{ \mathbf{z}_i \in \mathcal{D}_s \ \middle\vert \ p \leq \frac{1}{n_s} \hspace{-1em}\sum_{\substack{\mathbf{z}_j \sim \hat{\mathbb{P}}_{|\mathcal{D}_s|}^{-\mathbf{z}_i}: \\ j\in [\![  n_s ]\!]}} \hspace{-1em}\Psi_{\|\cdot\|, t,s}(i,j,\epsilon),  \forall i \in [\![  |\mathcal{D}_s| ]\!] \right\},
\end{equation*}
where $n_s = \frac{|\mathcal{D}_s|}{|\mathcal{D}|} n$ and $\epsilon_s = \frac{|\mathcal{D}_s|}{|\mathcal{D}|} \epsilon$. From this, we assemble an approximation of our original outlier set:
\begin{equation}\tag{\texttt{sSWAD}}\label{eq:swod_split}
    \mathcal{O} \approx \bigcup_{s=1}^S\mathcal{O}_s.
\end{equation}

\subsection{Fast Euclidian approximation}
We now present another approximation for when computational efficiency is the main requirement, e.g., large datasets, limited computational power, and a low number of threads. 

Our next proposition shows that the transportation plan between a distribution minus a sample and the same distribution minus another sample can be roughly approximated by the norm between the two removed samples. 
Consider the order-$t$ Wasserstein distance between two $N$-sample empirical distributions, $\hat{\mathbb{U}}_N$ and $\hat{\mathbb{V}}_N$, with samples denoted $\mathbf{u}_i$ and $\mathbf{v}_i$, respectively, $\forall i \in [\![  N ]\!]$:
\begin{equation}
    W_{\|\cdot\|,t}(\hat{\mathbb{U}}_N, \hat{\mathbb{V}}_N) = \inf_{\pi} \left(\frac{1}{N} \sum_{i=1}^{N} \|\mathbf{u}_i - \mathbf{v}_{\pi(i)}\|^t\right)^{1/t},
\end{equation}
over $\pi$, all possible permutations of $N$ elements and for $t\geq 1$. In this setting, there are $N!$ possible transportation plans. In our case, we consider two distributions which only differ by a single sample, viz., $\hat{\mathbb{P}}_{N-1}^{-\mathbf{z}_k}$ and $\hat{\mathbb{P}}_{N-1}^{-\mathbf{z}_l}$ for $l, k \in \mathbb{N}$ such that $k \neq l$. We can visualize the two distributions as scatter plots. Only one difference should be present between the two: in one plot, $\mathbf{z}_k$ appears and $\mathbf{z}_l$ is missing; in the latter, we see the opposite situation. The easiest transportation plan to conceive, between each plot, is to send $\mathbf{z}_l$ to $\mathbf{z}_k$ without moving other samples. We denote this suboptimal single-sample transportation plan as $\tilde{W}$. The plan $\tilde{W}$ is as follows:
\begin{align}
    W_{\|\cdot\|,t}(\hat{\mathbb{P}}_{N-1}^{-\mathbf{z}_k}, \hat{\mathbb{P}}_{N-1}^{-\mathbf{z}_l}) &\leq \tilde{W}_{\|\cdot\|,t}(\hat{\mathbb{P}}_{N-1}^{-\mathbf{z}_k}, \hat{\mathbb{P}}_{N-1}^{-\mathbf{z}_l}) \notag \\
    &\hspace{-0.1cm}= \sqrt[t]{\frac{1}{N-1} \left(0 + \ldots + 0 +   \|\mathbf{z}_k - \mathbf{z}_{l}\|^t\right)}\notag\\
    &\hspace{-0.1cm}= \frac{\|\mathbf{z}_k - \mathbf{z}_{l}\|}{\sqrt[t]{(N-1)}} , \label{eq:suboptimal_transport}
\end{align}
for $l, k \in [\![  N ]\!]$ such that $k \neq l$. Note that $\tilde{W}$ is generally suboptimal because the power of norms do not respect the triangle inequality for $t>1$ \cite{carlen2021inequalities}. Thus, a more cost-effective transport plan could involve the displacement of more than one sample. Nevertheless, it is an intuitive and computationally efficient transportation. In the special case where $t=1$, we can show that the inequality becomes an equality. 
\begin{lemma}\label{lem:single_sample_transport}
    For $t=1$, the following inequality holds: $${W}_{\|\cdot\|, 1}(\hat{\mathbb{P}}_{N-1}^{-\mathbf{z}_k}, \hat{\mathbb{P}}_{N-1}^{-\mathbf{z}_l}) = \tilde{W}_{\|\cdot\|, 1}(\hat{\mathbb{P}}_{N-1}^{-\mathbf{z}_k}, \hat{\mathbb{P}}_{N-1}^{-\mathbf{z}_l}),$$
    for all $l, k \in [\![  N ]\!] :k \neq l$.
\end{lemma}
\textit{Proof:}
Denote by $\mathbf{z}_{j_{i}}$ any intermediate samples in the transport of $\mathbf{z}_l$ to $\mathbf{z}_k$ such that ${j_{i}}\neq l \neq k \quad \forall l, k, j_{i} \in [\![  N ]\!]$ and $\forall i \in [\![  N-1 ]\!]$, where index $i$ is used to differentiate distinct intermediate samples. Using this notation and the triangle inequality, we can show that:
\begin{equation}
\begin{aligned}
        \| \mathbf{z}_k - \mathbf{z}_l\|&= \|\mathbf{z}_k - \mathbf{z}_{j_{1}} + \mathbf{z}_{j_{1}} - \mathbf{z}_l\| \\
        &\leq \|\mathbf{z}_k - \mathbf{z}_{j_{1}}\| + \|\mathbf{z}_{j_{1}} - \mathbf{z}_l\|\\
        & \leq \|\mathbf{z}_k - \mathbf{z}_{j_{1}}\| +   \| \mathbf{z}_{j_{1}} - \mathbf{z}_{j_{2}}\| + \|\mathbf{z}_{j_{2}} - \mathbf{z}_l\| \\
        &\leq \ldots
\end{aligned}\label{eq:triangle_l1}
\end{equation}
which means that for $t=1$, the single sample transport is less than or equal to any other permutation involving the transport of more than one sample. 

From (\ref{eq:triangle_l1}), we conclude that $\tilde{W}_{\|\cdot\|, 1}(\hat{\mathbb{P}}_{N-1}^{-\mathbf{z}_k}, \hat{\mathbb{P}}_{N-1}^{-\mathbf{z}_l}) = {W}_{\|\cdot\|, 1}(\hat{\mathbb{P}}_{N-1}^{-\mathbf{z}_k}, \hat{\mathbb{P}}_{N-1}^{-\mathbf{z}_l}) $ and thus it follows from (\ref{eq:suboptimal_transport}) that $\tilde{W}_{\|\cdot\|, t>1}(\hat{\mathbb{P}}_{N-1}^{-\mathbf{z}_k}, \hat{\mathbb{P}}_{N-1}^{-\mathbf{z}_l}) \geq {W}_{\|\cdot\|, t>1}(\hat{\mathbb{P}}_{N-1}^{-\mathbf{z}_k}, \hat{\mathbb{P}}_{N-1}^{-\mathbf{z}_l}) $. \hfill $\square$ 

To continue, we will need the following lemma.
\begin{lemma}[Remark 6.6 \cite{villani2008optimal}]\label{lem:villani_inequality}
    For $1\leq t  \leq q < +\infty$, the following inequality stands: $$W_{\|\cdot\|, t} \leq W_{\|\cdot\|, q}.$$
\end{lemma}
We now bound ${W}_{\|\cdot\|, t}(\hat{\mathbb{P}}_{N-1}^{-\mathbf{z}_k}, \hat{\mathbb{P}}_{N-1}^{-\mathbf{z}_l}) $.

\begin{proposition}\label{prop:wasserstein_inequality}
    For $1\leq t  < \infty$ and $\forall l, k \in [\![  N ]\!] :k \neq l$ , we obtain the following bounds: \begin{equation}
        \frac{\|\mathbf{z}_k - \mathbf{z}_{l}\|}{{(N-1)}} \leq {W}_{\|\cdot\|, t}(\hat{\mathbb{P}}_{N-1}^{-\mathbf{z}_k}, \hat{\mathbb{P}}_{N-1}^{-\mathbf{z}_l}) \leq \frac{\|\mathbf{z}_k - \mathbf{z}_{l}\|}{\sqrt[t]{(N-1)}}
    \end{equation}
\end{proposition}
\textit{Proof:} The proof follows from combining Lemmata~\ref{lem:single_sample_transport} and~\ref{lem:villani_inequality} as well as (\ref{eq:suboptimal_transport}). \hfill $\square$

\begin{figure}[!tb]
    \centering
    \includegraphics[width=1\linewidth]{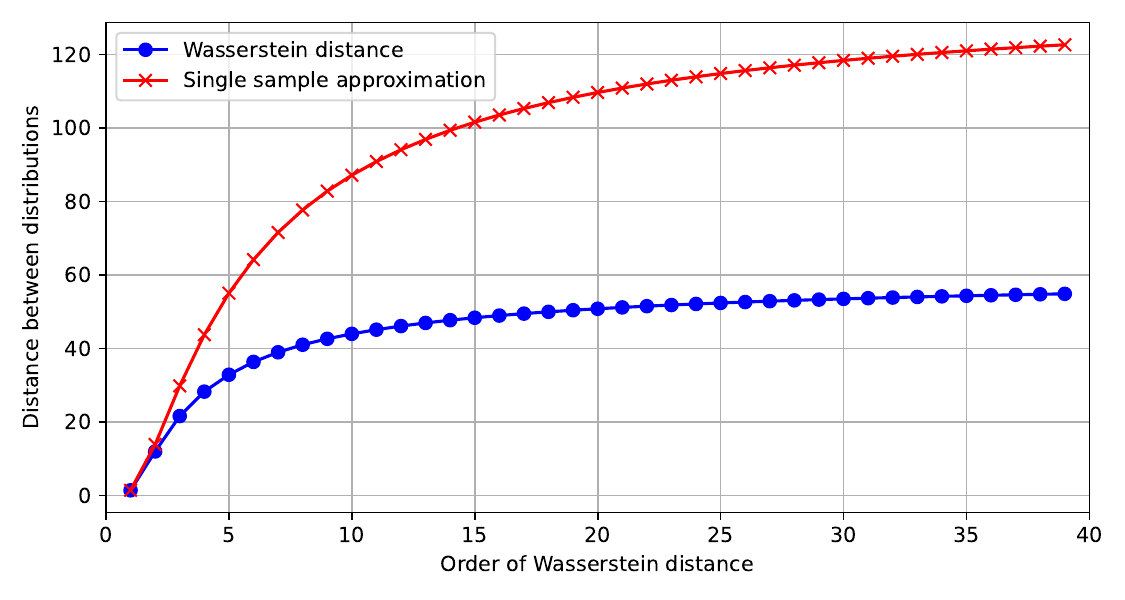}
    \caption{Empirical illustration of Proposition~\ref{prop:wasserstein_inequality} for $\|\cdot\|_2$}
    \label{fig:test_wasserstein_exactness}
\end{figure}

We present, in Figure~\ref{fig:test_wasserstein_exactness}, a visual representation of Proposition~\ref{prop:wasserstein_inequality} on an empirical measure of 100 samples generated with a Gaussian distribution. As can be seen, the property holds.
As such, based on Proposition~\ref{prop:wasserstein_inequality}, we introduce our fast Euclidian AD (\texttt{FEAD}) method for $W_{\|\cdot\|_2,t}$:
\begin{equation}\label{eq:fead}\tag{\texttt{FEAD}}
    \mathcal{O} \approx \left\{ \mathbf{z}_i \in \mathcal{D} \middle\vert \ p \leq \frac{1}{n} \hspace{-1.25em}\sum_{\substack{\mathbf{z}_j \sim \hat{\mathbb{P}}_{N-1}^{-\mathbf{z}_i}:\\ j\in [\![  n ]\!]}}\hspace{-1em}\mathbbm{1}\left( \frac{\|\mathbf{z}_i - \mathbf{z}_j\|_2}{\sqrt[t]{N-1}} \geq \eta \right), \forall i \in [\![  N ]\!] \right\}\hspace{-0.25em},
\end{equation} 
where $\eta$ is the threshold of the Euclidian distance. This method might not be as accurate as~\ref{eq:swod} to filter out-of-sample data points when $t\neq 1$, but it is extremely fast and has a direct relation with the Wasserstein distance. Because~\ref{eq:swod} and \ref{eq:fead} use the same principle, they share similar classification patterns when $\epsilon$ and $\eta$ are tuned accordingly, as can be seen in Figure~\ref{fig:2d_toy} of Appendix~\ref{app:qualitative}.

\section{Case study}

We now illustrate our methods in several experiments.

\subsection{Qualitative study}\label{sec:swfilter_qualitative}
\begin{figure}[!tb]
     \centering
     \begin{subfigure}{0.45\linewidth}
         \includegraphics[width=\linewidth]{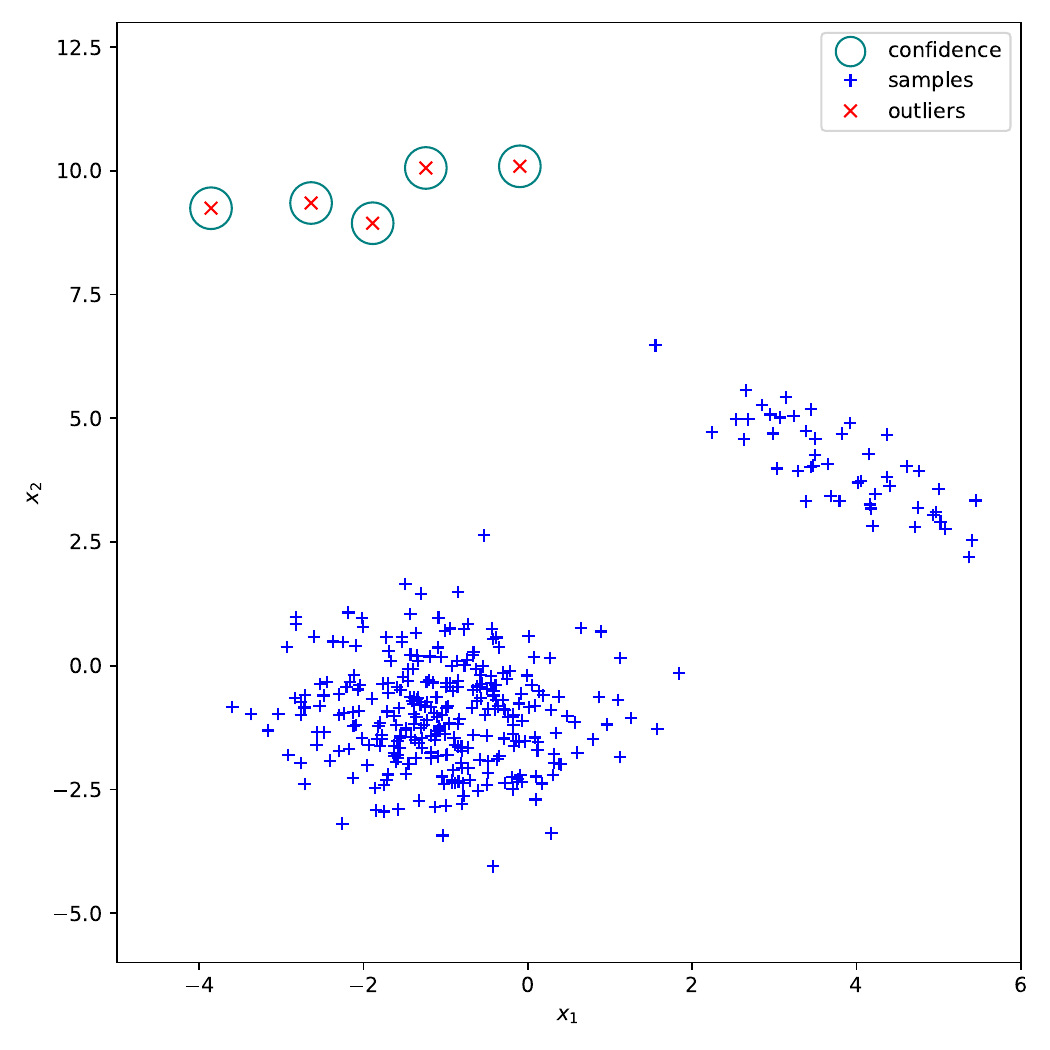}
         \caption{$\epsilon=0.1$}
     \end{subfigure}
     \hfill
     \begin{subfigure}{0.45\linewidth}
         \includegraphics[width=\linewidth]{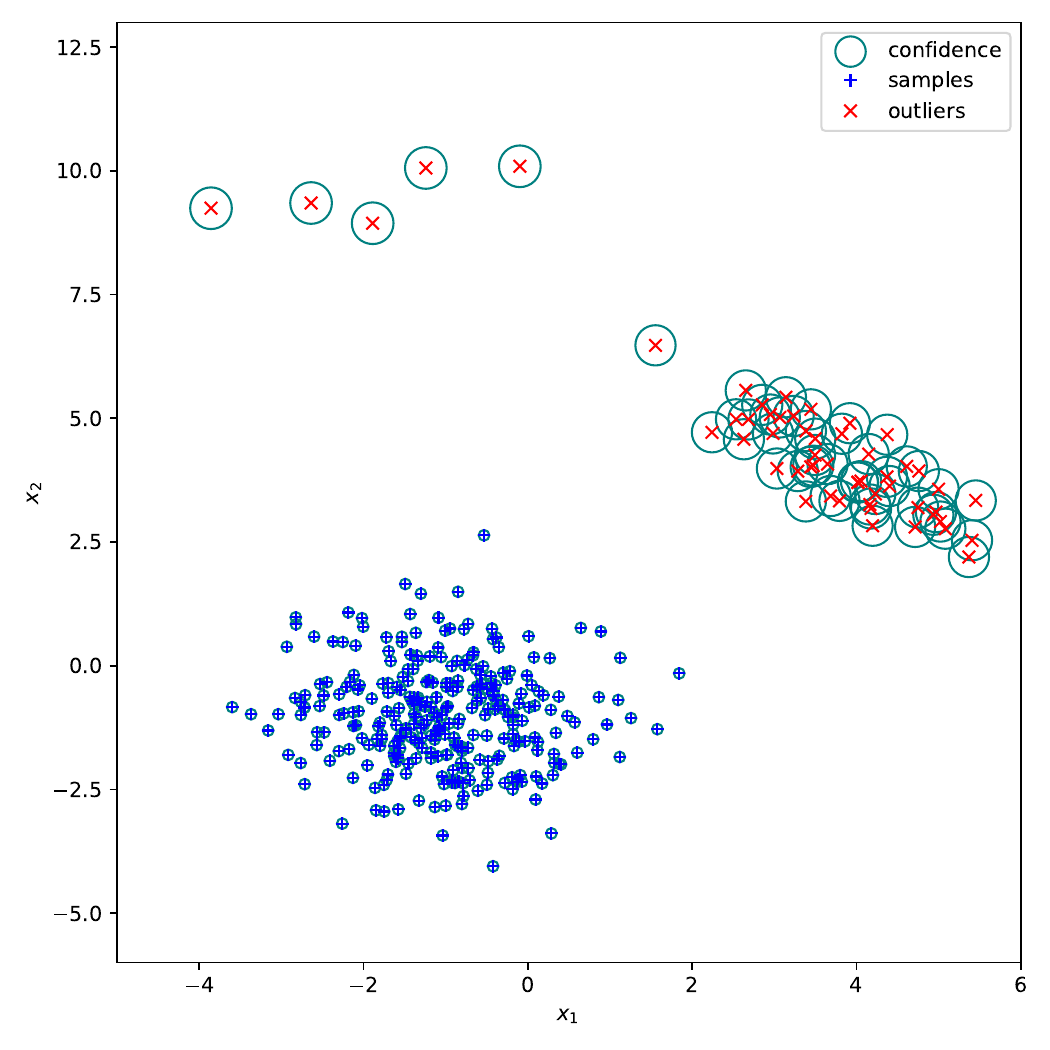}
         \caption{$\epsilon=0.05$}
     \end{subfigure}
     \hfill
         \begin{subfigure}{0.45\linewidth}
         \includegraphics[width=\linewidth]{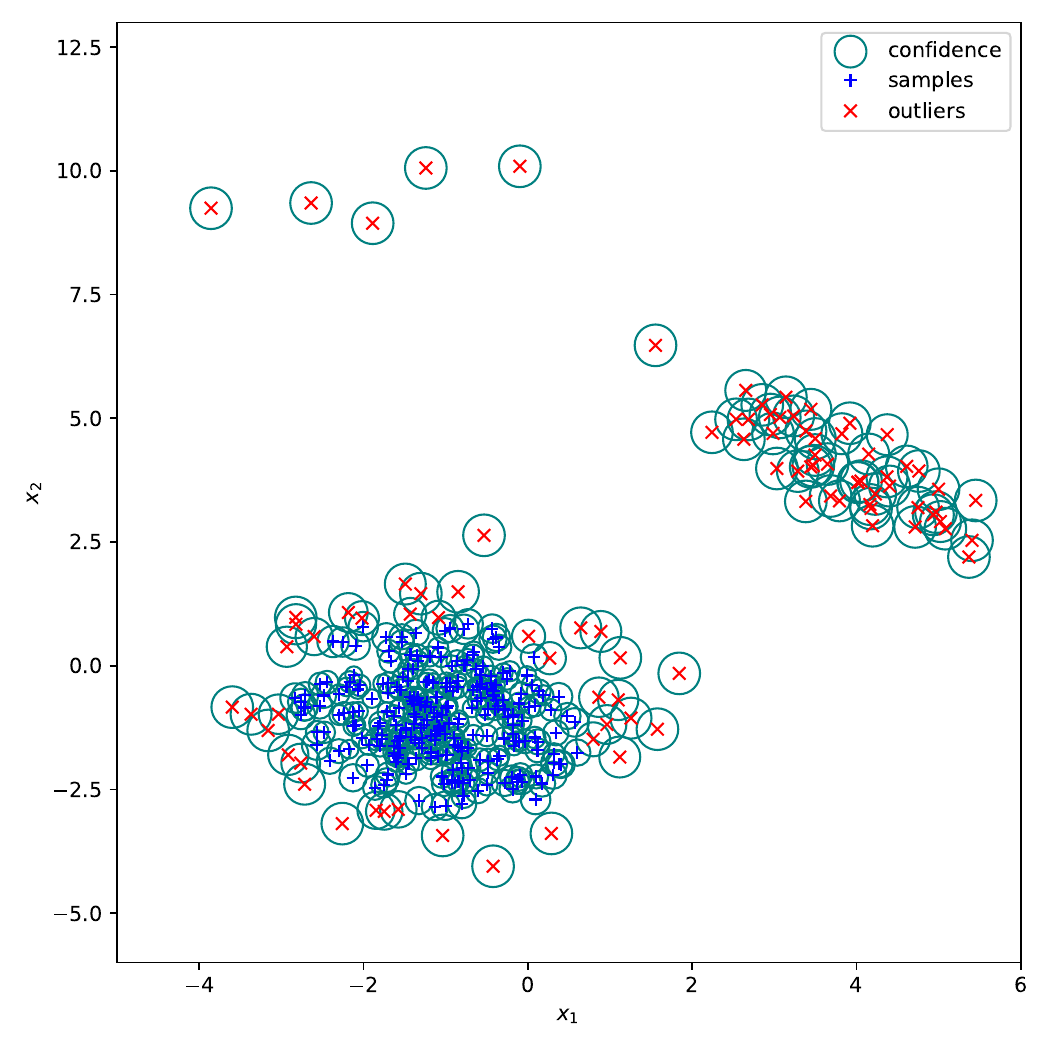}
         \caption{$\epsilon=0.01$}
     \end{subfigure}
     \caption{Labelling of the SW filter for different values of $\epsilon$}
     \label{fig:2d_tut}
 \end{figure}

In this section we show the AD mechanism of the SW filter on a small two-dimensional example. We generate three Gaussian distributions with different population sizes: the first distribution is the majority group, the second is the minority group, and the third represents clear statistical outliers.
We merge the three distributions into a single dataset to test our method. We vary the threshold $\epsilon$ to see how the filter behaves qualitatively. The results are presented in Figure~\ref{fig:2d_tut}.
We observe that we can generate three filtering scenarios by modifying the value of $\epsilon$. With $\epsilon = 0.1$, we filter only the statistical outliers. With $\epsilon = 0.05$, we only keep the majority group. And, with $\epsilon = 0.01$, we only keep the samples closest to the barycentre of the majority group. This is very interesting in a safe ML pipeline as we can tune the \textit{conservatism} of the training dataset that is used at each run during hyperparameter optimization. See Appendix~\ref{app:qualitative} for a qualitative comparison with other AD models.
 \subsection{Numerical experiments}\label{sec:swfilter_experiments}

We now test our method for training data selection. We use regression datasets found on the UCI ML repository \cite{markelleUCI}. To keep computation requirements low, we compare~\ref{eq:fead},~\ref{eq:swod_split}, LOF, Isolation Forest, and no filtering. A numerical comparison including~\ref{eq:swod} is provided in Appendix~\ref{app:qualitative}.
For each dataset, 10\% of the data is kept for the testing phase. From the remaining 90\%, we use 30\% of it during validation and train on the rest. Both training and validation samples are corrupted by a Gaussian noise applied after data scaling. The Gaussian noise is centred at the origin with a variance of 0.05 on all features and the label. In each trial, we train a non-regularized shallow convex neural network (SCNN)~\cite{pilanci2020neural, mishkin2022scnn} constructed with a maximum of 25 neurons in the hidden layer. The SCNN is trained using the $\ell_1$-loss function on the data filtered by each AD method. We allow for 100 trials per filtering method per dataset and use the tree-structured Parzen estimator for hyperparameter optimization~\cite{bergstra2011algorithms}. The hyperparameter search space is available in Appendix~\ref{app:data_sel}. 
The datasets comprise \texttt{wine}~\cite{wine_quality_186}, \texttt{forest} \cite{forest_fires_162}, \texttt{solar}~\cite{solar_flare_89}, \texttt{energy}~\cite{energy_efficiency_242}, \texttt{concrete}~\cite{concrete_compressive_strength_165}, and \texttt{stock}~\cite{istanbul_stock_exchange_247}.

\begin{figure}[!tb]
     \centering
     \begin{subfigure}{1\linewidth}
         \includegraphics[width=0.95\linewidth]{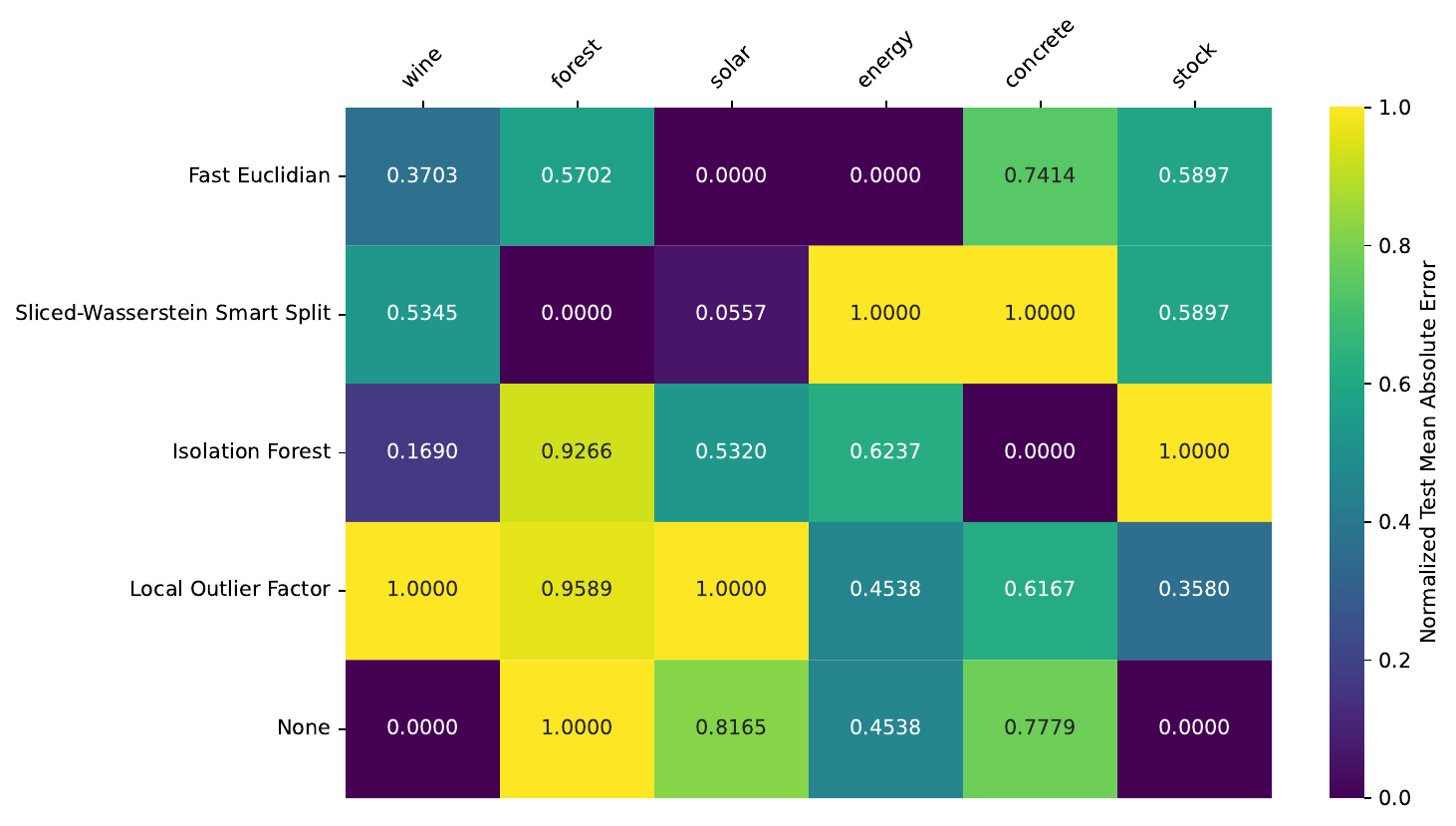}
         \caption{Normalized MAE in testing}
     \end{subfigure}
     \hfill
     \begin{subfigure}{1\linewidth}
         \includegraphics[width=0.95\linewidth]{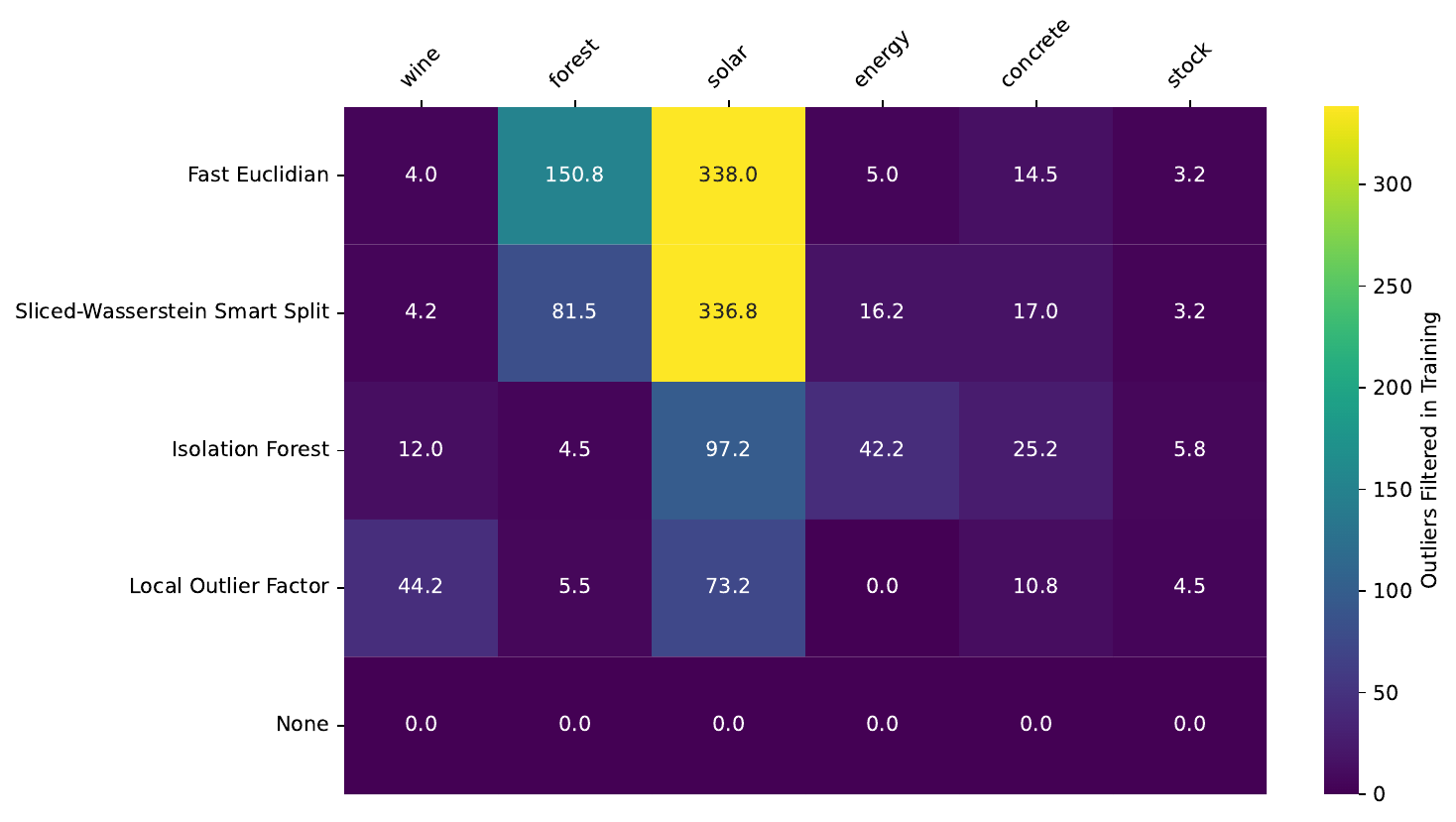}
         \caption{Average number of samples filtered in the training set}
     \end{subfigure}
     \caption{Results of the data selection experiment}
     \label{fig:MAE_dataselect}
 \end{figure}

We repeat each experiment four times on different testing splits and extract the average results. The normalized average mean absolute error (MAE) obtained in testing for the best trials in validation are displayed in Figure~\ref{fig:MAE_dataselect}, where we also provide the average number of training samples filtered by each method. In each column, the highest MAE is equal to~1, and the lowest is equal to~0. 
We also provide the absolute errors in Figure~\ref{fig:data_select_absolute} from Appendix~\ref{app:data_sel}.

 From Figure~\ref{fig:MAE_dataselect}, we make the following observations. While no method completely outperforms the others, we remark that~\ref{eq:fead} never has the highest error compared to other methods as they have at least one column equal to 1. Surprisingly, the no-filtering method has the lowest error on two datasets: we believe that the use of the $\ell_1$-loss and the decision to train SCNNs may have helped to avoid overfitting during training and validation. Nevertheless, we observe that \ref{eq:fead} has by far the lowest average normalized errors over the six datasets (\ref{eq:fead}: 0.3786;~\ref{eq:swod_split}: 0.5300; {Isolation Forest}: 0.5419; {Local Outlier Factor}: 0.7312; {None}: 0.5081).

\section{Open-source dataset}\label{sec:open_source}
\begin{figure}[!htb]
     \centering
     \includegraphics[width=0.95\linewidth, trim={0.25cm 0 0.2cm 0.2cm}, clip]{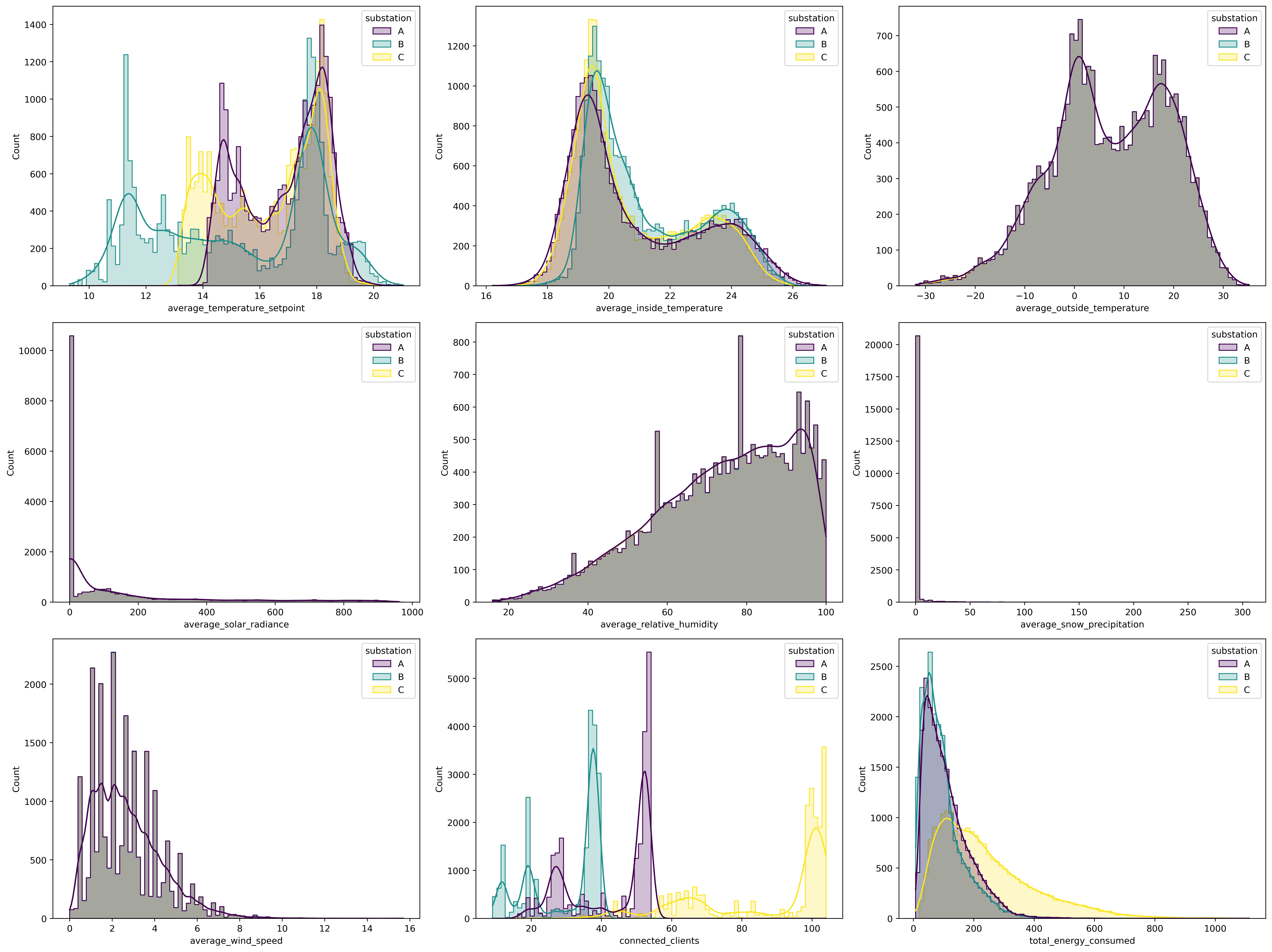}
     \caption{Distribution of key features for each substation}
     \label{fig:dist}
 \end{figure}

The dataset we share contains the aggregated hourly consumption of 197 anonymous LCPR testers located in three neighbouring substations in Montréal, Canada. The dataset was developed in collaboration with Hydro-Québec with the data from Hilo. Additional hourly weather data and LCPR information are also provided. 
We remark that outliers and anomalies are present in the dataset because of metering and telemetry issues or even blackouts, e.g., an aberrant (and impossible) 32.2 MWh energy consumption is registered at some point. Figure~\ref{fig:dist} shows the distribution count of some key features for each substation. We observe that meteorological features are identical for each substation as they are geographically adjacent and share the same meteorological station. 
We refer readers to Appendix~\ref{app:data} 
for the full list of features available in the dataset as well as additional analyses and visualizations of the dataset's features and labels.

We now propose a first benchmark on our LCPR dataset. Our goal is to predict the aggregated hourly consumption at each substation during winter when peak demand is critical. Following the literature~\cite{weng2015GP}, we propose a simple yet meaningful benchmark and implement a Gaussian process~\cite{williams1995gaussian} in a rolling horizon fashion. 
The Gaussian process's kernel combines a radial basis function kernel, an exponential sine squared kernel to model periodicity, and a white noise kernel.
Samples dating from before 2023-12-15 are used for hyperparameter tuning, while those between 2023-12-15 and 2024-04-15, during the most recent winter CPR season, are used for testing. We train a model per week and the training window size is a hyperparameter to be tuned. Because of its good performance in the data selection experiment of Section~\ref{sec:swfilter_experiments} and its low computational needs, we use \ref{eq:fead} for data selection during hyperparameter optimization.

Our method is interesting because it can filter clear out-of-sample points while hopefully avoiding sparse LCPR events that other methods could consider as local outliers. We refer interested readers to our GitHub page for the full overview of the hyperparameter search space and the construction of the kernel used in GP training. 
 
 \begin{table}[h]
 \renewcommand{\arraystretch}{1.2}
     \centering
     \caption{Absolute test errors of the best validation run for each substation}
     \begin{tabular}{c|ccc}
     \hline 

     \hline 
       \textbf{Substation} & \textbf{A} & \textbf{B} & \textbf{C}\\ \hline

       \textbf{MAE [kWh]} & 20.49103 & 17.90663 & 41.21746 \\

       \textbf{RMSE [kWh]} & 26.08270 & 22.39355  & 51.25044\\
     \hline 

     \hline 
     \end{tabular}
     \label{tab:bench}
 \end{table}

Testing MAE and root mean squared errors (RMSEs) are presented in Table~\ref{tab:bench} for each substation.
We remark that the errors for Substation~C are higher than for Substations A and B, but the value of its label is also higher. During testing, Substations A and B record consumption between approximately 50 kWh and 300 kWh, compared to 100 kWh and 600 kWh for Substation C. 
See Appendix~\ref{app:data} to visualize the test predictions.

\section{Closing remarks}\label{sec:swfilter_closing}
In this work, we propose an anomaly detection method for training data selection based on the SW distance. To keep computational tractability when using larger datasets, we propose two approximations: (i) a split method in which we decompose the original dataset in multiple reduced-cardinality representations (splits) and (ii) a fast method based on the Euclidian distance. We remark that method (i) is especially interesting when multi-threading is available. 

We observe that our proposition is adequate for filtering global outliers yet sometimes fails at detecting local outliers. We argue that this is quite interesting to robustify adversarial training datasets. We insist that, like most similar methods
our proposition is not a \textit{jack-of-all-trades}, but it is relevant for specific kinds of data corruption, relies on a proven mathematical distance, is unsupervised, and can easily be included in the hyperparameter optimization phase of an ML pipeline. One of the biggest limitations of this method is the lack of proper theoretical guarantees for out-of-sample model predictive performance. This is a topic for future work.

We also propose a new open-source dataset on LCPR and provide a first forecasting benchmark using our methodology for data selection. 

\section*{Acknowledgement} 
We wish to highlight the help of Christophe Bélanger, Dr.~Salma Naccache, and Dr.~Bertrand Scherrer for the opening and the validation of the dataset. 
\bibliographystyle{ieeetr} 
\bibliography{ref.bib}

\begin{thebibliography}{10}

\bibitem{pudjianto2007virtual}
D.~Pudjianto, C.~Ramsay, and G.~Strbac, ``{Virtual power plant and system integration of distributed energy resources},'' {\em {IET Renewable Power Generation}}, vol.~1, no.~1, pp.~10--16, 2007.

\bibitem{zhang2019comprehensive}
G.~Zhang, C.~Jiang, and X.~Wang, ``{Comprehensive review on structure and operation of virtual power plant in electrical system},'' {\em IET Generation, Transmission \& Distribution}, vol.~13, no.~2, pp.~145--156, 2019.

\bibitem{albadi2007DR}
M.~H. Albadi and E.~F. El-Saadany, ``{Demand Response in Electricity Markets: An Overview},'' in {\em 2007 IEEE Power Engineering Society General Meeting}, pp.~1--5, 2007.

\bibitem{Lijesen_2007_realtime_elasticity}
M.~G. Lijesen, ``The real-time price elasticity of electricity,'' {\em Energy Economics}, vol.~29, no.~2, pp.~249--258, 2007.

\bibitem{Yang2013_ToU}
P.~Yang, G.~Tang, and A.~Nehorai, ``A game-theoretic approach for optimal time-of-use electricity pricing,'' {\em IEEE Transactions on Power Systems}, vol.~28, no.~2, pp.~884--892, 2013.

\bibitem{6254817}
Q.~Zhang and J.~Li, ``{Demand response in electricity markets: A review},'' in {\em 2012 9th International Conference on the European Energy Market}, pp.~1--8, 2012.

\bibitem{liu2020data}
C.~Liu, P.~Nitschke, S.~P. Williams, and D.~Zowghi, ``{Data quality and the Internet of Things},'' {\em Computing}, vol.~102, no.~2, pp.~573--599, 2020.

\bibitem{bonnotte2013unidimensional}
N.~Bonnotte, {\em Unidimensional and evolution methods for optimal transportation}.
\newblock PhD thesis, Universit{\'e} Paris Sud-Paris XI; Scuola normale superiore (Pise, Italie), 2013.

\bibitem{chatzivasileiadis2022machine}
S.~Chatzivasileiadis, A.~Venzke, J.~Stiasny, and G.~Misyris, ``{Machine learning in power systems: Is it time to trust it?},'' {\em IEEE Power and Energy Magazine}, vol.~20, no.~3, pp.~32--41, 2022.

\bibitem{goldstein2016AD}
M.~Goldstein and S.~Uchida, ``{A Comparative Evaluation of Unsupervised Anomaly Detection Algorithms for Multivariate Data},'' {\em PLOS ONE}, vol.~11, pp.~1--31, 04 2016.

\bibitem{Breuning2000LOF}
M.~M. Breunig, H.-P. Kriegel, R.~T. Ng, and J.~Sander, ``{LOF: identifying density-based local outliers},'' {\em Proceedings of the 2000 ACM SIGMOD International Conference on Management of Data}, p.~93–104, 2000.

\bibitem{Liu2008IsolationForest}
F.~T. Liu, K.~M. Ting, and Z.-H. Zhou, ``{Isolation Forest},'' in {\em 2008 Eighth IEEE International Conference on Data Mining}, pp.~413--422, 2008.

\bibitem{angiulli2002knn}
F.~Angiulli and C.~Pizzuti, ``{Fast Outlier Detection in High Dimensional Spaces},'' {\em Proceedings of the Sixth European Conference on the Principles of Data Mining and Knowledge Discovery}, vol.~2431, pp.~15--26, 08 2002.

\bibitem{Tang2002CLOF}
J.~Tang, Z.~Chen, A.~W.-C. Fu, and D.~W.-L. Cheung, ``{Enhancing Effectiveness of Outlier Detections for Low Density Patterns},'' in {\em Proceedings of the 6th Pacific-Asia Conference on Advances in Knowledge Discovery and Data Mining}, PAKDD '02, (Berlin, Heidelberg), p.~535–548, Springer-Verlag, 2002.

\bibitem{Amer2013OneClassSVM}
M.~Amer, M.~Goldstein, and S.~Abdennadher, ``{Enhancing one-class support vector machines for unsupervised anomaly detection},'' {\em Proceedings of the ACM SIGKDD Workshop on Outlier Detection and Description}, p.~8–15, 2013.

\bibitem{ducoffe2019anomaly}
M.~Ducoffe, I.~Haloui, and J.~S. Gupta, ``{Anomaly detection on time series with Wasserstein GAN applied to PHM},'' {\em International Journal of Prognostics and Health Management}, vol.~10, no.~4, 2019.

\bibitem{wang2024WOOD}
Y.~Wang, W.~Sun, J.~Jin, Z.~Kong, and X.~Yue, ``{WOOD: Wasserstein-Based Out-of-Distribution Detection},'' {\em {IEEE Transactions on Pattern Analysis and Machine Intelligence}}, vol.~46, no.~2, pp.~944--956, 2024.

\bibitem{mercado2014enabling}
A.~Mercado, R.~Mitchell, S.~Earni, R.~Diamond, and E.~Alschuler, ``{Enabling interoperability through a common language for building performance data},'' {\em Proceedings of the 2014 ACEEE Summer Study on Energy Efficiency in Buildings}, 2014.

\bibitem{feng2024gridedge}
F.~Li, I.~Kocar, and A.~Lesage-Landry, ``{A Rapid Method for Impact Analysis of Grid-Edge Technologies on Power Distribution Networks},'' {\em IEEE Transactions on Power Systems}, vol.~39, no.~1, pp.~1530--1542, 2024.

\bibitem{Venzke2021Worstcase}
A.~Venzke and S.~Chatzivasileiadis, ``Verification of neural network behaviour: Formal guarantees for power system applications,'' {\em IEEE Transactions on Smart Grid}, vol.~12, no.~1, pp.~383--397, 2021.

\bibitem{Chen2020_DRO}
R.~Chen and I.~C. Paschalidis, ``Distributionally robust learning,'' {\em Foundations and Trends{\textregistered} in Optimization}, vol.~4, no.~1-2, pp.~1--243, 2020.

\bibitem{Misyris2020PINN_PES}
G.~S. Misyris, A.~Venzke, and S.~Chatzivasileiadis, ``Physics-informed neural networks for power systems,'' in {\em 2020 IEEE Power \& Energy Society General Meeting (PESGM)}, pp.~1--5, 2020.

\bibitem{Jospine2022_handson_BNN}
L.~V. Jospin, H.~Laga, F.~Boussaid, W.~Buntine, and M.~Bennamoun, ``Hands-on bayesian neural networks—a tutorial for deep learning users,'' {\em IEEE Computational Intelligence Magazine}, vol.~17, no.~2, pp.~29--48, 2022.

\bibitem{pallage2024wasserstein}
J.~Pallage and A.~Lesage-Landry, ``{Wasserstein Distributionally Robust Shallow Convex Neural Networks},'' {\em arXiv preprint arXiv:2407.16800}, 2024.

\bibitem{pallage2024sliced}
J.~Pallage, B.~Scherrer, S.~Naccache, C.~Bélanger, and A.~Lesage-Landry, ``{Sliced-Wasserstein-based Anomaly Detection and Open Dataset for Localized Critical Peak Rebates},'' in {\em {NeurIPS 2024 Workshop on Tackling Climate Change with Machine Learning}}, 2024.

\bibitem{ruan2024data}
G.~Ruan, D.~Qiu, S.~Sivaranjani, A.~S. Awad, and G.~Strbac, ``Data-driven energy management of virtual power plants: A review,'' {\em Advances in Applied Energy}, p.~100170, 2024.

\bibitem{reis2022fundamentals}
J.~Reis and M.~Housley, {\em Fundamentals of data engineering}.
\newblock O'Reilly Media, Inc., 2022.

\bibitem{kleppmann2017designing}
M.~Kleppmann, {\em {Designing data-intensive applications: The big ideas behind reliable, scalable, and maintainable systems}}.
\newblock O'Reilly Media, Inc., 2017.

\bibitem{panaretos2019statistical}
V.~M. Panaretos and Y.~Zemel, ``{Statistical aspects of Wasserstein distances},'' {\em Annual Review of Statistics and its Application}, vol.~6, no.~1, pp.~405--431, 2019.

\bibitem{kolouri2019generalized}
S.~Kolouri, K.~Nadjahi, U.~Simsekli, R.~Badeau, and G.~Rohde, ``{Generalized sliced Wasserstein distances},'' {\em Advances in Neural Information Processing Systems}, vol.~32, 2019.

\bibitem{carlen2021inequalities}
E.~A. Carlen, R.~L. Frank, P.~Ivanisvili, and E.~H. Lieb, ``{Inequalities for Lp-Norms that Sharpen the Triangle Inequality and Complement Hanner’s Inequality},'' {\em The Journal of geometric analysis}, vol.~31, pp.~4051--4073, 2021.

\bibitem{villani2008optimal}
C.~Villani, {\em Optimal transport: old and new}, vol.~338.
\newblock Springer, 2008.

\bibitem{markelleUCI}
M.~Kelly, R.~Longjohn, and K.~Nottingham, ``{UCI Machine Learning Repository},'' 2023.

\bibitem{pilanci2020neural}
M.~Pilanci and T.~Ergen, ``{Neural networks are convex regularizers: Exact polynomial-time convex optimization formulations for two-layer networks},'' in {\em International Conference on Machine Learning}, pp.~7695--7705, PMLR, 2020.

\bibitem{mishkin2022scnn}
A.~Mishkin, A.~Sahiner, and M.~Pilanci, ``{Fast Convex Optimization for Two-Layer ReLU Networks: Equivalent Model Classes and Cone Decompositions},'' in {\em International Conference on Machine Learning}, pp.~15770--15816, PMLR, 2022.

\bibitem{bergstra2011algorithms}
J.~Bergstra, R.~Bardenet, Y.~Bengio, and B.~K{\'e}gl, ``Algorithms for hyper-parameter optimization,'' vol.~24, pp.~1--9, 2011.

\bibitem{wine_quality_186}
P.~Cortez, A.~Cerdeira, F.~Almeida, T.~Matos, and J.~Reis, ``{Wine Quality}.'' UCI Machine Learning Repository, 2009.

\bibitem{forest_fires_162}
P.~Cortez and A.~Morais, ``{Forest Fires}.'' UCI Machine Learning Repository, 2007.

\bibitem{solar_flare_89}
G.~Brashaw, ``{Solar Flare}.'' UCI Machine Learning Repository, 1989.

\bibitem{energy_efficiency_242}
A.~Tsanas and A.~Xifara, ``{Energy Efficiency}.'' UCI Machine Learning Repository, 2012.

\bibitem{concrete_compressive_strength_165}
I.-C. Yeh, ``{Concrete Compressive Strength}.'' UCI Machine Learning Repository, 1998.

\bibitem{istanbul_stock_exchange_247}
O.~Akbilgic, ``{Istanbul stock exchange}.'' UCI Machine Learning Repository, 2013.

\bibitem{weng2015GP}
Y.~Weng and R.~Rajagopal, ``{Probabilistic baseline estimation via Gaussian process},'' in {\em {2015 IEEE Power \& Energy Society General Meeting}}, pp.~1--5, 2015.

\bibitem{williams1995gaussian}
C.~Williams and C.~Rasmussen, ``Gaussian processes for regression,'' {\em Advances in Neural Information Processing Systems}, vol.~8, 1995.

\bibitem{scikit-learn}
F.~Pedregosa, G.~Varoquaux, A.~Gramfort, V.~Michel, B.~Thirion, O.~Grisel, M.~Blondel, P.~Prettenhofer, R.~Weiss, V.~Dubourg, J.~Vanderplas, A.~Passos, D.~Cournapeau, M.~Brucher, M.~Perrot, and E.~Duchesnay, ``Scikit-learn: Machine learning in {P}ython,'' {\em Journal of Machine Learning Research}, vol.~12, pp.~2825--2830, 2011.

\bibitem{campos2016evaluation}
G.~O. Campos, A.~Zimek, J.~Sander, R.~J. Campello, B.~Micenkov{\'a}, E.~Schubert, I.~Assent, and M.~E. Houle, ``{On the evaluation of unsupervised outlier detection: measures, datasets, and an empirical study},'' {\em Data Mining and Knowledge Discovery}, vol.~30, pp.~891--927, 2016.

\bibitem{bergstra2015hyperopt}
J.~Bergstra, B.~Komer, C.~Eliasmith, D.~Yamins, and D.~D. Cox, ``{Hyperopt: a Python library for model selection and hyperparameter optimization},'' {\em Computational Science \& Discovery}, vol.~8, no.~1, p.~014008, 2015.

\bibitem{spearman04}
C.~Spearman, ``{The proof and measurement of association between two things},'' {\em American Journal of Psychology}, vol.~15, pp.~88--103, 1904.

\bibitem{chen2016xgboost}
T.~Chen and C.~Guestrin, ``{XGBoost: A Scalable Tree Boosting System},'' in {\em Proceedings of the 22nd ACM SIGKDD International Conference on Knowledge Discovery and Data Mining}, KDD '16, (New York, NY, USA), p.~785–794, Association for Computing Machinery, 2016.

\bibitem{lundberg2017SHAP}
S.~M. Lundberg and S.-I. Lee, ``A unified approach to interpreting model predictions,'' in {\em Proceedings of the 31st International Conference on Neural Information Processing Systems}, NIPS'17, (Red Hook, NY, USA), p.~4768–4777, Curran Associates Inc., 2017.

\end{thebibliography}

\appendix

We last provide supplementary material on SW distance-based data selection.


\subsection{Supplementary qualitative study}\label{app:qualitative}
\begin{figure*}[tb]
     \centering
     \includegraphics[width=\linewidth]{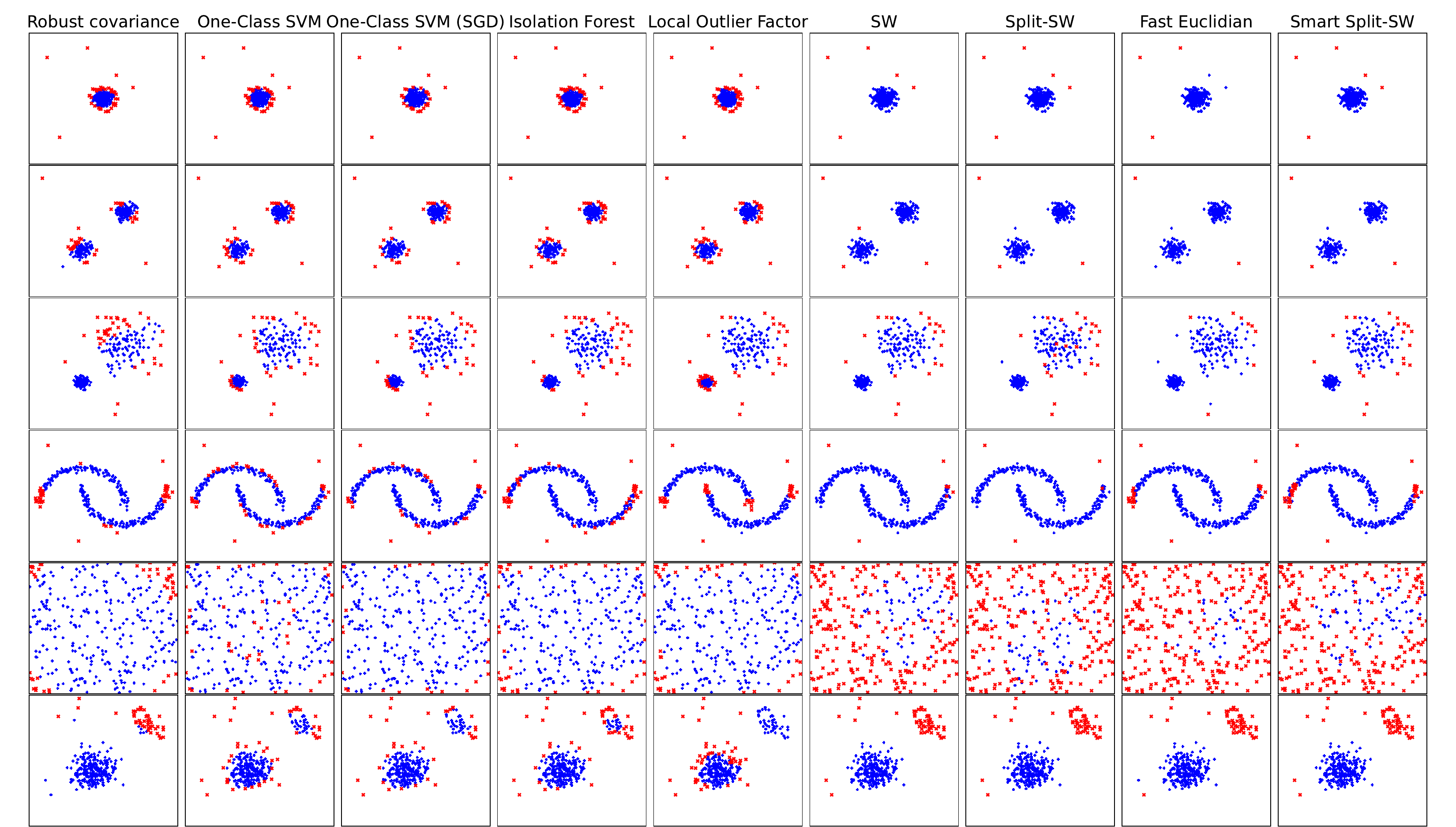}
     \caption{AD comparison on multiple synthetic datasets}
     \label{fig:2d_toy}
 \end{figure*}

To complement~Section~\ref{sec:swfilter_qualitative}, we compare in Figure~\ref{fig:2d_toy} different AD algorithms on synthetic datasets provided in \texttt{scikit-learn}'s example collection~\cite{scikit-learn}. Recall that most of these algorithms were introduced in Section~\ref{sec:intro} except for Robust Covariance, which is a method originating from \texttt{scikit-learn} \cite{scikit-learn}. Hyperparameters are fixed for each algorithm to see how a single hyperparameter choice influences the labelling on each dataset. 

As can be seen in Figure~\ref{fig:2d_toy}, \texttt{SWAD} and \texttt{FEAD} are better at isolating outliers when there is a clear majority group but are not as precise in identifying local outliers based on local density like, e.g., one-class SVM.

\subsection{Supplementary numerical study for anomaly detection}
\label{app:num_ad}

We run a thorough AD experiment with commonly used real-world benchmark datasets. We run experiments on the \texttt{Lymphography}, \texttt{Ionosphere}, \texttt{Glass}, \texttt{Shuttle}, \texttt{WPBC}, \texttt{Arrhythmia}, and \texttt{Pima}, datasets presented in~\cite{campos2016evaluation} for AD benchmarking. We compare isolation forest and LOF to order-$2$~\ref{eq:swod} with respect to $\|\cdot\|_2$ and~\ref{eq:fead}. We omit~\ref{eq:swod_split} because the size of the experiment permits the use of~\ref{eq:swod}. For this study, the performance indicators are defined as follows:
\begin{align*}
A &= \frac{T_\text{p} + T_\text{n}}{T_\text{p} + F_\text{p} + T_\text{n} + F_\text{n}}\\
P &= \frac{T_\text{p}}{T_\text{p} + F_\text{p}},    
\end{align*}
where $T_\text{p}$, $F_\text{p}$, $T_\text{n}$, and $F_\text{n}$ stand for true positives, false positives, true negatives, and false negatives, respectively. We implement a grid search with~\texttt{hyperopt}~\cite{bergstra2015hyperopt} and select each model's run with the best accuracy $A$. We then extract the precision score $P$ from this run.
 
\begin{figure}[!htb]
     \centering
     \begin{subfigure}{1\linewidth}
     \centering
         \includegraphics[width=0.75\linewidth]{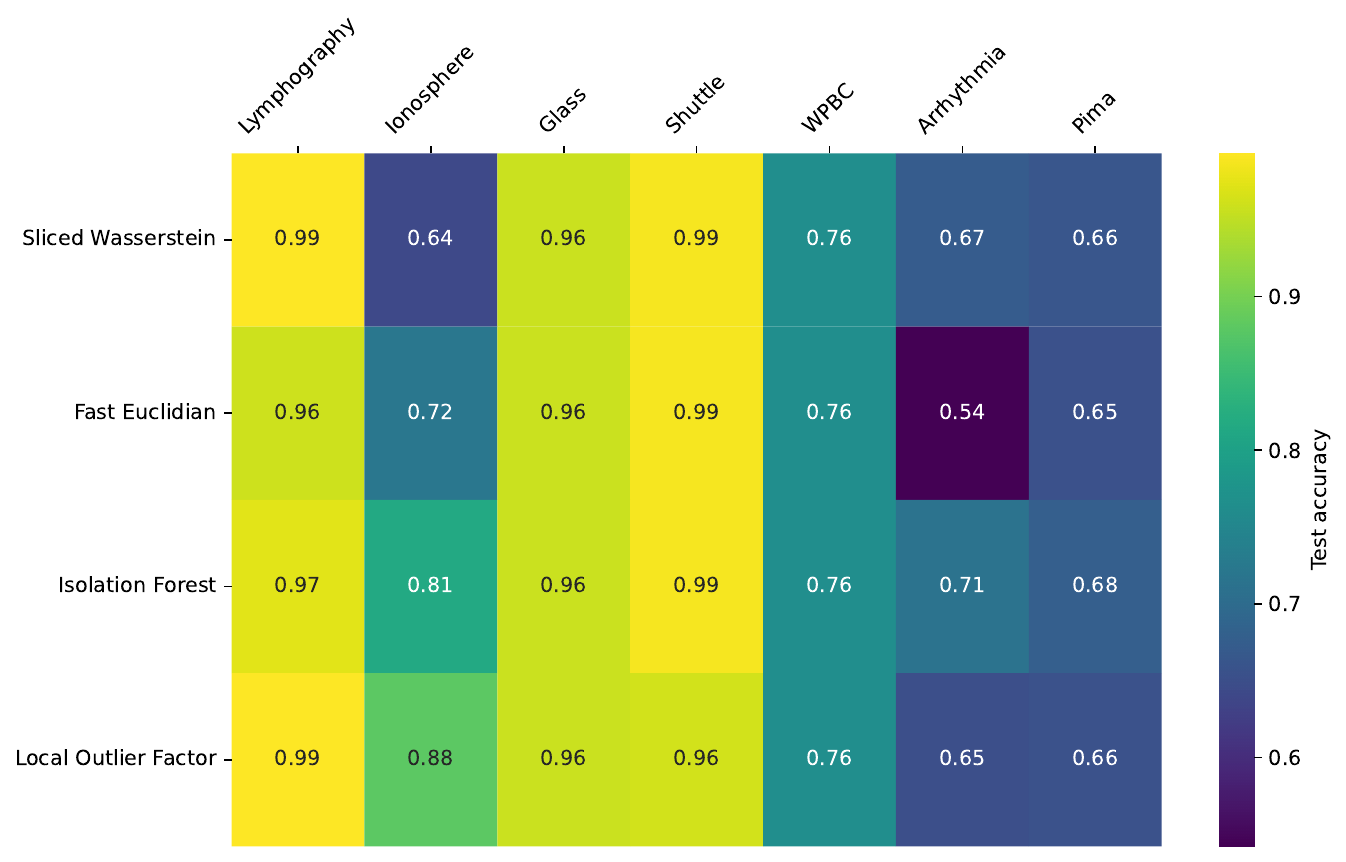}
         \caption{Accuracy $A$}
     \end{subfigure}
     \hfill
     \begin{subfigure}{1\linewidth}
     \centering
         \includegraphics[width=0.75\linewidth]{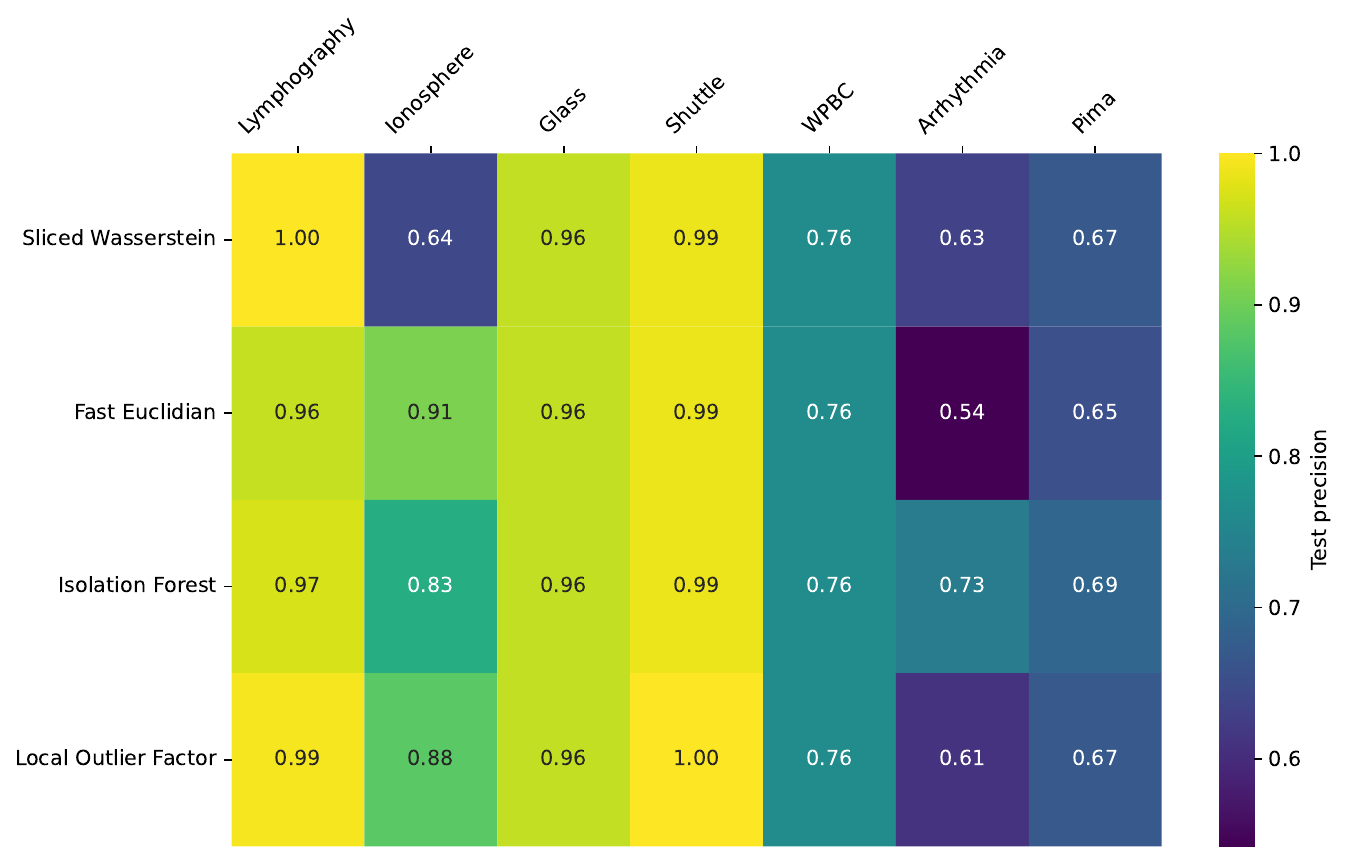}
         \caption{Precision $P$}
     \end{subfigure}
     \caption{Grid search for each AD model on each dataset}
     \label{fig:grid}
 \end{figure}

In all our tests, we use the order-$2$ sliced-Wasserstein distance for our method. The results are presented in Figure~\ref{fig:grid}. We observe a similar performance between each model, except on the \texttt{Ionosphere} dataset where the \texttt{SWAD} lags behind the other models, and on \texttt{Arrhythmia} where \texttt{FEAD} is underperforming. The SW filter's strength is that it considers the global distributional properties of the population to guide its labelling. We remark that our method fails at detecting local outliers as it is purely designed to locate global outliers that might be \textit{risky} or adversarial to keep in the training set. This is a reasonable choice to make when designing data preprocessing methods for safer ML training, but not necessarily to filter local outliers. Every experiment is available on our GitHub page. 
The hyperparameter intervals used in this study are presented in Table~\ref{tab:anomaly_detection_hyperparameters}. Because we do a grid search, the search space is discretized.  

 \begin{table}[tb]
 \centering
 \caption{Hyperparameters for the AD experiments}
 \resizebox{\linewidth}{!}{
 \begin{tblr}{colspec={m{2.5cm}|lll}, hline{1,6} = {1.5pt}, hline{2,3,4,5}={1pt}} 
 \textbf{Model} & \textbf{Hyperparameter} & \textbf{Description} & \textbf{Possible values} \\

 LOF & 
 {\texttt{n\_neighbors} \\ 
 \texttt{algorithm} \\ 
 \texttt{leaf\_size} \\ 
 \texttt{metric}}  & 
 {Number of neighbours \\ 
 Algorithm  \\ 
 Leaf size \\ 
 Distance metric} & 
 {$\{10, 25, 50, 75, 100, 500\}$ \\ 
 \{'auto', 'ball\_tree', 'kd\_tree', 'brute'\} \\ 
 $\{5, 10, 25, 50, 75, 100, 500\}$ \\ 
 \{'euclidean', 'manhattan', 'chebyshev', 'minkowski'\}} \\

~\ref{eq:swod} & 
 {\texttt{eps} \\ 
 \texttt{n} \\ 
 \texttt{n\_projections} \\ 
 \texttt{p}}  & 
 {Distance threshold \\ 
 Number of neighbours \\ 
 Number of projections \\ 
 Voting threshold} & 
 {$\{0.00001, 0.0001, 0.001, 0.01, 0.1, 1, 10, 100, 1000\}$ \\ 
 $\{25, 75\}$ \\ 
 $\{100, 200\}$ \\ 
 $\{0.5, 0.7, 0.9\}$} \\

~\ref{eq:fead} & 
 {\texttt{eps} \\ 
 \texttt{n} \\ 
 \texttt{p}}  & 
 {Distance threshold \\ 
 Number of neighbours \\ 
 Voting threshold} & 
 {$\{0.00001, 0.0001, 0.001, 0.01, 0.1, 1, 10, 100, 1000\}$ \\ 
 $\{25, 75, 150, 200, 500\}$ \\ 
 $\{0.5, 0.7, 0.9\}$} \\
 Isolation Forest & 
 {\texttt{n\_estimators} \\ 
 \texttt{max\_samples} \\ 
 \texttt{contamination} \\ 
 \texttt{max\_features}}  & 
 {Number of base estimators \\ 
 Number of samples \\ 
 Expected proportion of outliers \\ 
 Number of features} & 
 {$\{10, 50, 100, 500\}$ \\ 
 \{1, $n/4$, $n/2$, $3n/4$, $n$\} \\ 
 $\{0.01, 0.15, 0.3, 0.45\}$ \\ 
 $\{1, \frac{d}{2}, d\}$} \\
 \end{tblr}}
 \label{tab:anomaly_detection_hyperparameters}
 \end{table}

 \subsection{Supplementary content to the data selection experiment}\label{app:data_sel}
\begin{figure}[hb]
     \centering
     \includegraphics[width=1\linewidth]{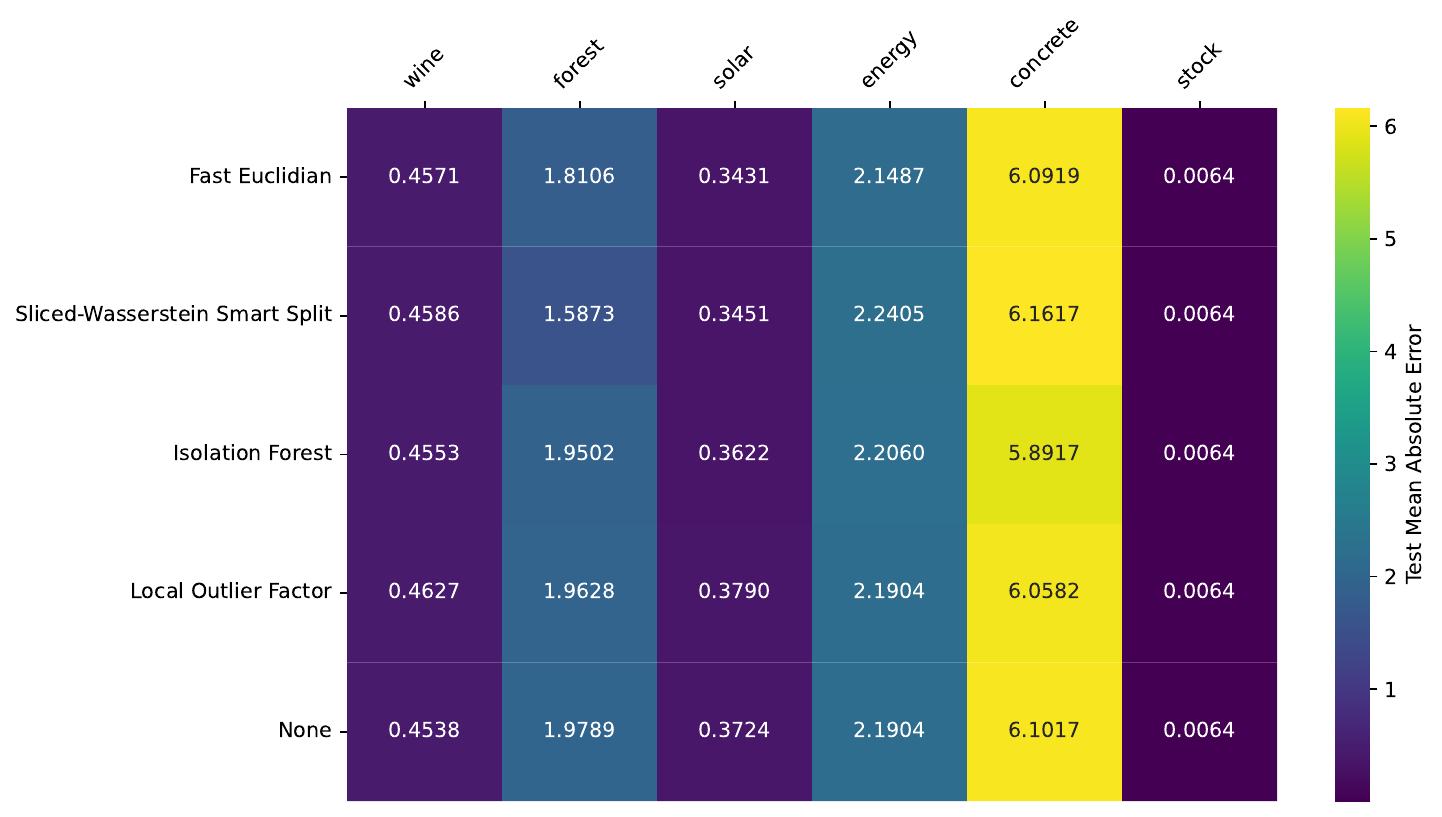}
     \caption{Absolute MAE in testing of the data selection experiment}
     \label{fig:data_select_absolute}
 \end{figure}

 To complement~Section~\ref{sec:swfilter_experiments}, we also provide the averaged absolute MAE values of the experiment in Figure~\ref{fig:data_select_absolute}.

Lastly, the hyperparameters involved in the experiment are listed together with their search interval in Table~\ref{tab:anomaly_hyperparameters}.

 \begin{table}[tb]
     \centering
     \caption{Hyperparameters for the data selection experiment}
     \resizebox{\linewidth}{!}{
     \begin{tblr}{colspec={m{2.5cm}|lll}, hline{1,6} = {1.5pt}, hline{2,3,4,5}={1pt}} 
     \textbf{Model} & \textbf{Hyperparameter} & \textbf{Description} & \textbf{Possible values} \\
   
     LOF & 
     {\texttt{n\_neighbors} \\ 
     \texttt{algorithm} \\ 
     \texttt{leaf\_size} \\ 
     \texttt{metric}}  & 
     {Number of neighbours \\ 
     Algorithm  \\ 
     Leaf size \\ 
     Distance metric} & 
     {$[5, 300]$ (integer) \\ 
     {\{'auto', 'ball\_tree', 'kd\_tree', 'brute'\}} \\ 
     $[5, 300]$ (integer) \\ 
     \{'euclidean', 'manhattan'\}} \\

    ~\ref{eq:swod_split} & 
     {\texttt{eps} \\ 
     \texttt{n} \\ 
     \texttt{n\_projections} \\ 
     \texttt{p} \\
     \texttt{n\_clusters} \\
     \texttt{n\_splits}}  & 
     {Distance threshold \\ 
     Number of neighbours \\ 
     Number of projections \\ 
     Voting threshold \\
     Number of clusters \\
     Number of splits} & 
     {$[\mathrm{e}^{-12}, \mathrm{e}^1]$ \\ 
     $\{150, 300\}$ \\ 
     $\{40\}$ \\ 
     $\{0.7, 0.8, 0.9\}$ \\
     $\{3, 4\}$ \\
     $\{3, 4\}$} \\
   
    ~\ref{eq:fead} & 
     {\texttt{eps} \\ 
     \texttt{n} \\ 
     \texttt{p}}  & 
     {Distance threshold \\ 
     Number of neighbours \\ 
     Voting threshold} & 
     {$[\mathrm{e}^{-12}, \mathrm{e}^7]$ \\ 
     $\{150, 300\}$ \\ 
     $\{0.7, 0.8, 0.9\}$} \\
    
     Isol. Forest & 
     {\texttt{n\_estimators} \\ 
     \texttt{max\_samples} \\ 
     \texttt{contamination} \\ 
     \texttt{max\_features}}  & 
     {Number of base estimators \\ 
     Proportion of samples \\ 
     Expected proportion of outliers \\ 
     Proportion of features} & 
     {$[5, 300]$ (integer) \\ 
     $(0, 1]$ \\ 
     $[\mathrm{e}^{-7}, \mathrm{e}^{-0.7}]$ \\ 
     $(0, 1]$} \\
     \end{tblr}}
     \label{tab:anomaly_hyperparameters}
 \end{table}

 \subsection{Detailed analysis of the LCPR dataset}\label{app:data}
 In this section, we present the full list of features and labels for the LCPR dataset as well as additional insights for future analysis. All columns from the datasets are detailed in Table~\ref{tab:features_opendataset}
 
\begin{table}[tb]
\caption{Description of features and label of the dataset}
\label{tab:features_opendataset}
\renewcommand{\arraystretch}{1.0}
    \centering
 \resizebox{\linewidth}{!}{
     \begin{tblr}[caption ={Description of features and label of the dataset},
         label = {tab:features_opendataset}]{colspec={l|m{5.5cm}|c},hline{1,25} = {1.5pt}, hline{2,3,4,5,6,7,8,9,10,11,12,13,14,15,16,17,18,19,20,21,22,23,24}={1pt}} 
     \textbf{Name} & \textbf{Description} & \textbf{Possible values}  \\
     \texttt{substation} & Substation identifier & \{\textquotesingle A\textquotesingle, \textquotesingle B\textquotesingle, \textquotesingle C\textquotesingle\}\\
     \texttt{timestamp\_local} & Timestamp in local time (UTC-5) and ISO 8601 format [AAAA-MM-DD hh:mm:ss]& $-$\\
     \texttt{connected\_clients} & Number of clients connected to the substation during the considered hour & $\{9,10,\ldots,104\}$ \\
     \texttt{connected\_smart\_tstats} & Number of smart thermostats connected to the substation during the considered hour & $\{59,60,\ldots,1278\}$\\
     \texttt{average\_inside\_temperature} & Hourly average indoor  temperature measured by smart thermostats in substation [$^\circ$C]& $[16.21,27.08]$ \\
     \texttt{average\_temperature\_setpoint} & Hourly average setpoint of smart thermostats in substation [$^\circ$C] & $[9.31, 21.03]$\\
     \texttt{average\_outside\_temperature} & Hourly average outside temperature at substation [$^\circ$C] & $[-32.0, 35.2]$ \\
     \texttt{average\_solar\_radiance} & Hourly average solar radiance at substation [W/m$^2$] & $[0,961]$ \\
     \texttt{average\_relative\_humidity} & Hourly average relative humidity at substation [\%] & [0,100]\\
     \texttt{average\_snow\_precipitation} & Hourly average amount of snow precipitation at substation [mm] & [0.0,306.0]\\
     \texttt{average\_wind\_speed} & Hourly average wind speed at substation [m/s] & [0, 15.68 ]\\
     \texttt{date} & Date [AAAA-MM-DD]& {[2022-01-01,\\ 2024-06-30]}\\
     \texttt{month} & Month &  $\{1, 2, \ldots, 12\}$ \\
     \texttt{day} & Day of the month& $\{1, 2, \ldots, 31\}$\\
     \texttt{day\_of\_week} & Day of the week with Sunday and Saturday being 1 and 7, respectively&  $\{1, 2, \ldots, 7\}$\\
     \texttt{hour} & Hour of the day&  $\{0, 1, \ldots, 23\}$\\
      \texttt{challenge\_type} & Type of challenge during the given hour & {\{\textquotesingle None\textquotesingle, \textquotesingle CPR\textquotesingle, \\ \textquotesingle LCPR\textquotesingle\}}\\
       \texttt{challenge\_flag} & Flag indicating hours in challenge&  $\{0, 1\}$ \\
       \texttt{pre\_post\_challenge\_flag} & Flag indicating hours in pre-challenge or post-challenge&  $\{0, 1\}$ \\
     \texttt{is\_weekend} & Flag indicating weekends & $\{0, 1\}$\\
      \texttt{is\_holiday} & Flag indicating Québec holidays  &  $\{0, 1\}$\\
       \texttt{weekend\_holiday} & Flag indicating whether a weekend or a holiday &  $\{0, 1\}$\\
       \texttt{total\_energy\_consumed } &Hourly energy consumption of the substation [kWh]  &  $[7.45, 32240.17]$\\
 \end{tblr}}

\end{table}

\begin{figure}[!htb]
     \centering
     \begin{subfigure}{1\linewidth}
         \includegraphics[width=0.9\linewidth]{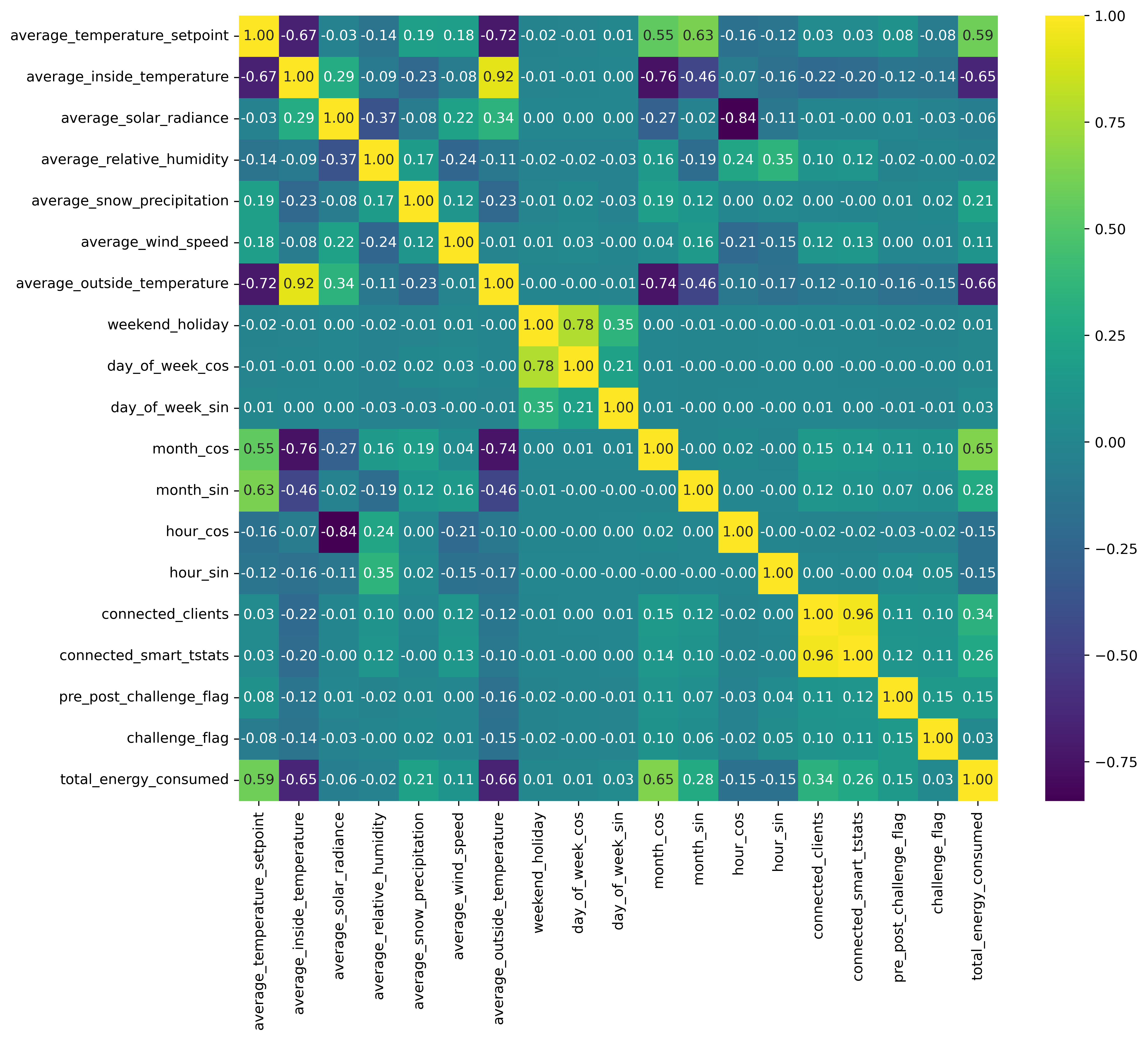}
         \vspace{-0.25cm}
         \caption{Substation A}\vspace{-0.25cm}
     \end{subfigure}
     \hfill
     
     \begin{subfigure}{1\linewidth}
         \includegraphics[width=0.9\linewidth]{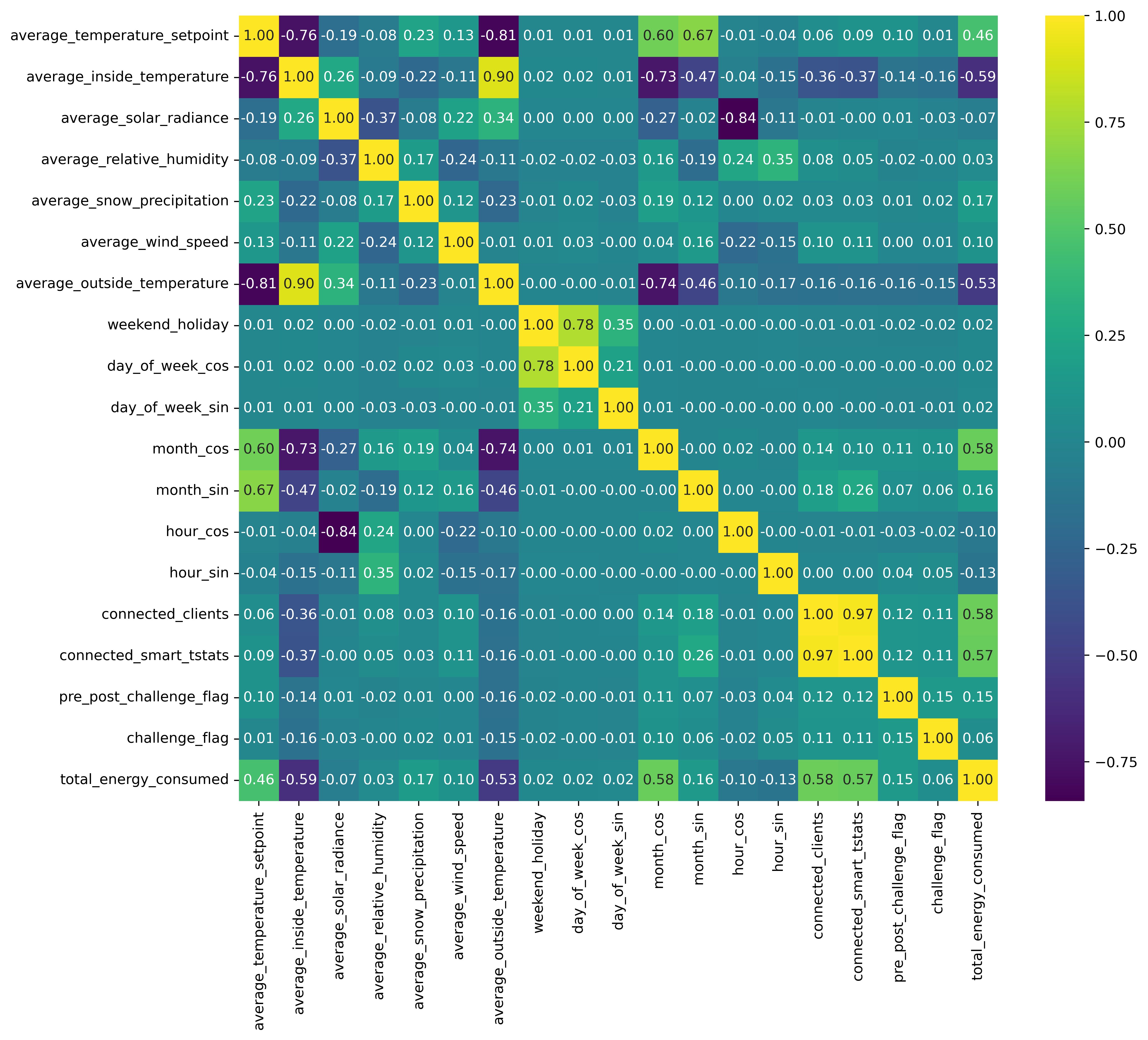}
         \vspace{-0.25cm}
         \caption{Substation B}
         \vspace{-0.25cm}
     \end{subfigure}
     \hfill
     
         \begin{subfigure}{1\linewidth}
         \includegraphics[width=0.9\linewidth]{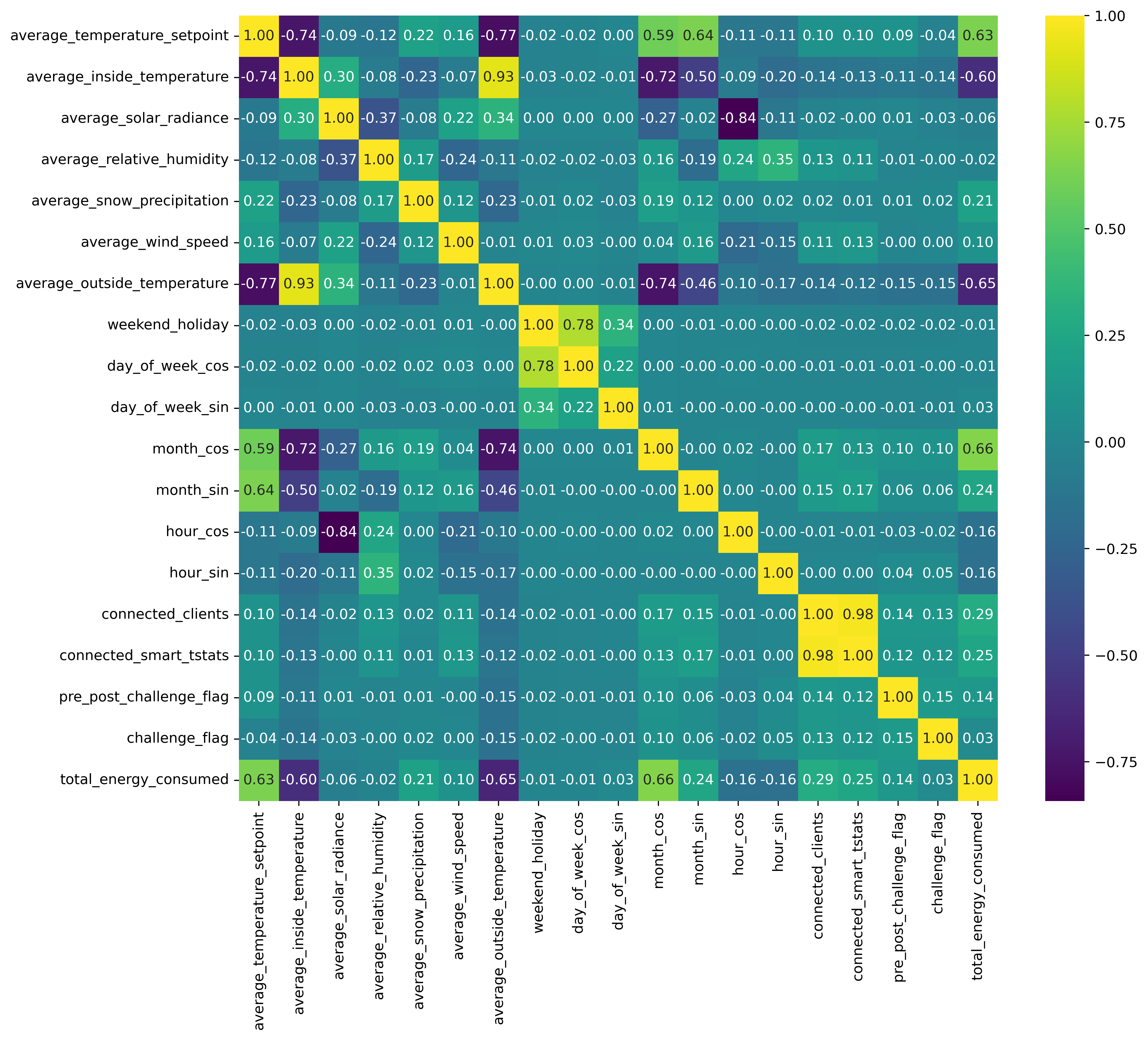}
         \vspace{-0.25cm}
         \caption{Substation C}
         \vspace{-0.25cm}
     \end{subfigure}
     \caption{Correlation heatmap of key features and label for each substation}
     \label{fig:corr}
 \end{figure}

\begin{figure}[!htb]
     \centering
     \includegraphics[width=0.75\linewidth]{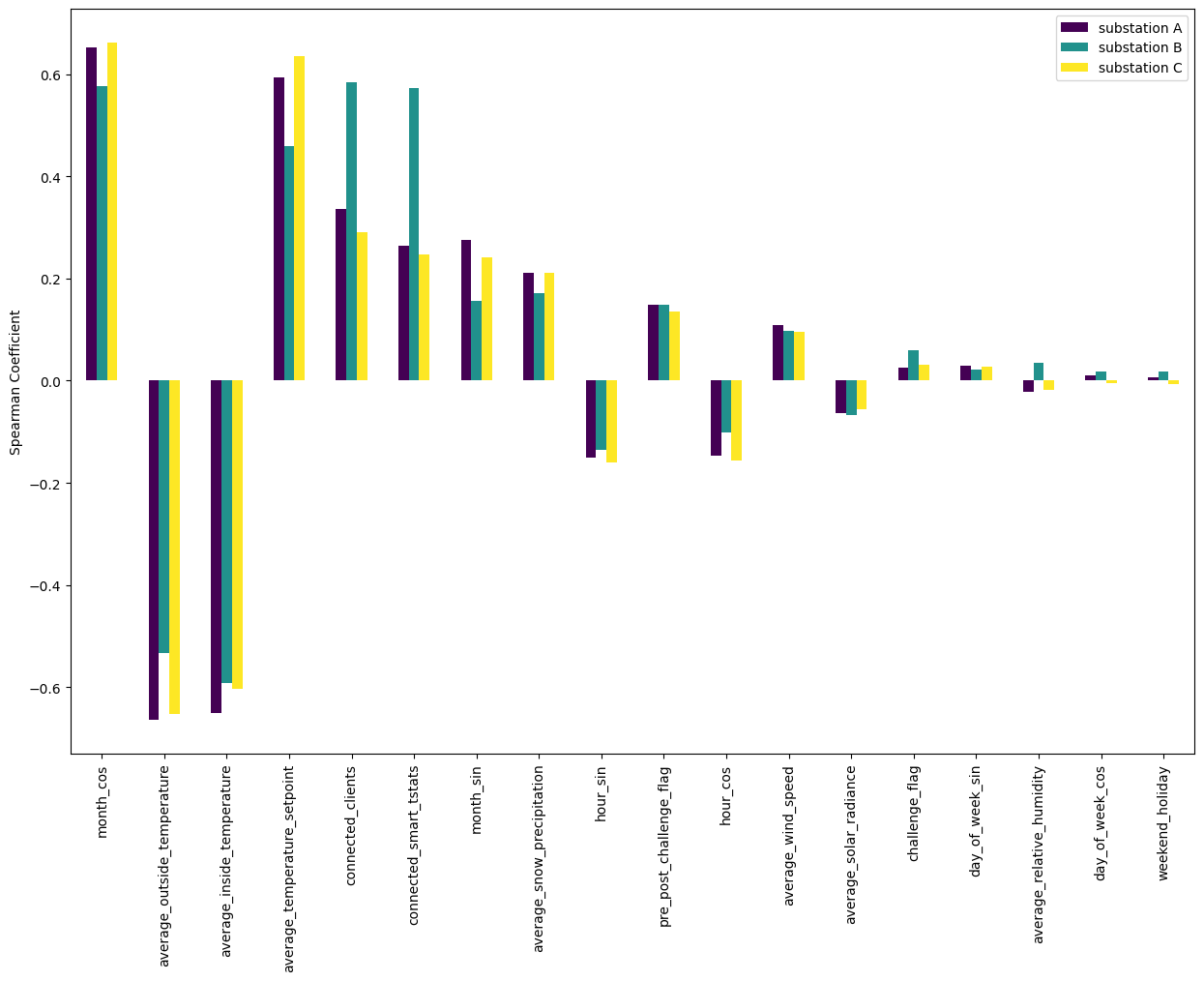}
     \caption{Spearman coefficients of key features for each substation}
     \label{fig:spear}
 \end{figure}

 Figure~\ref{fig:corr} shows a correlation heatmap of important features and label for each substation. We observe that each substation follows the same general tendencies. We notice significant correlations between the energy consumed, the month of the year, the outside temperatures, and the temperature setpoints. This is also highlighted in Figure~\ref{fig:spear} which presents the Spearman coefficients ranking~\cite{spearman04} between each feature and the label. A positive sign indicates that both the label and the feature increase or decrease in the same direction while a negative sign indicates an opposite direction. The coefficients are ranked in decreasing importance from left to right. 

\begin{figure}[tb]
     \centering
     \begin{subfigure}{1\linewidth}
         \centering\includegraphics[width=0.75\linewidth]{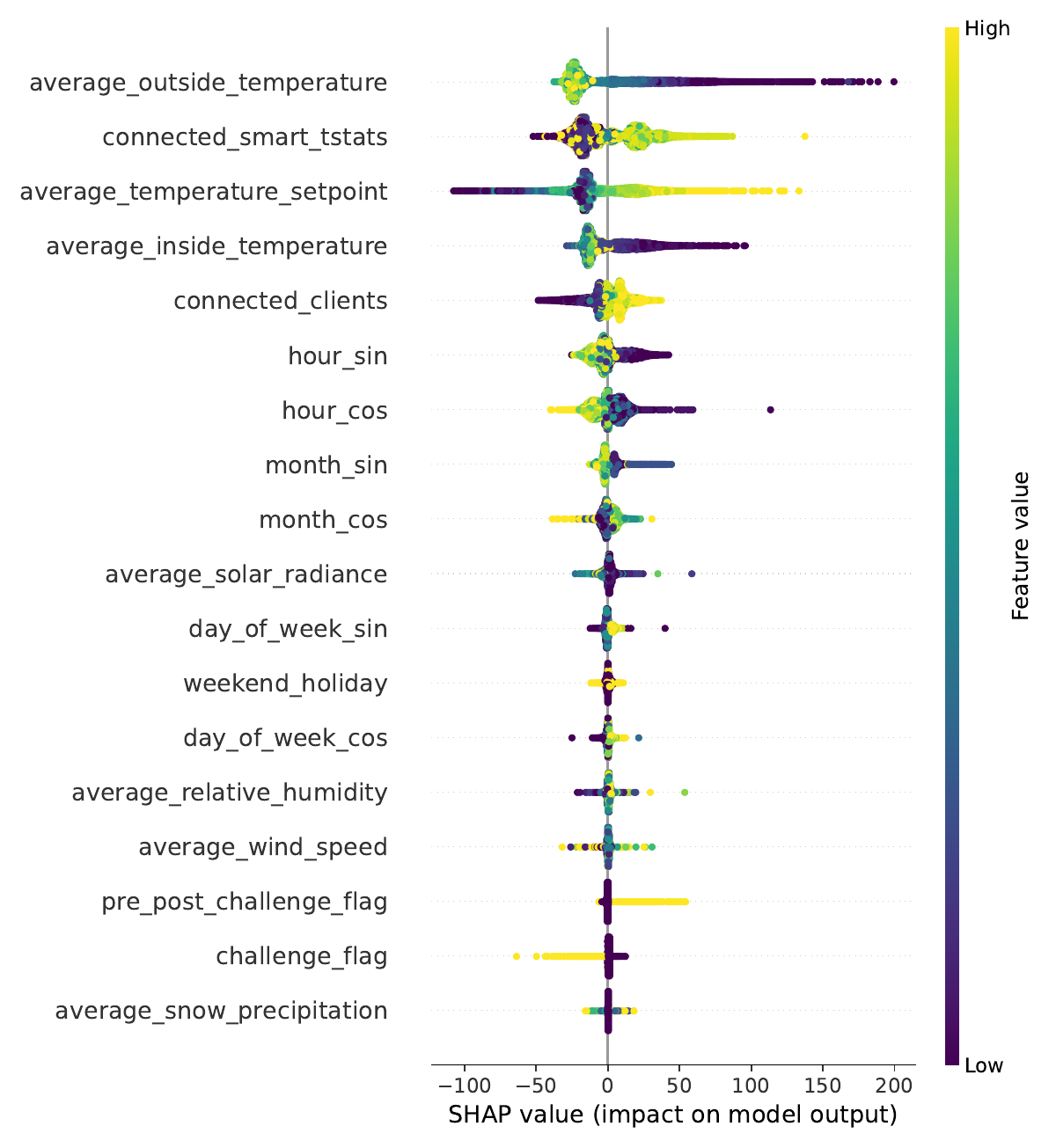}\vspace{-0.25cm}
         \caption{Substation A}\vspace{-0.25cm}
     \end{subfigure}
     \hfill
     
     \begin{subfigure}{1\linewidth}
         \centering\includegraphics[width=0.75\linewidth]{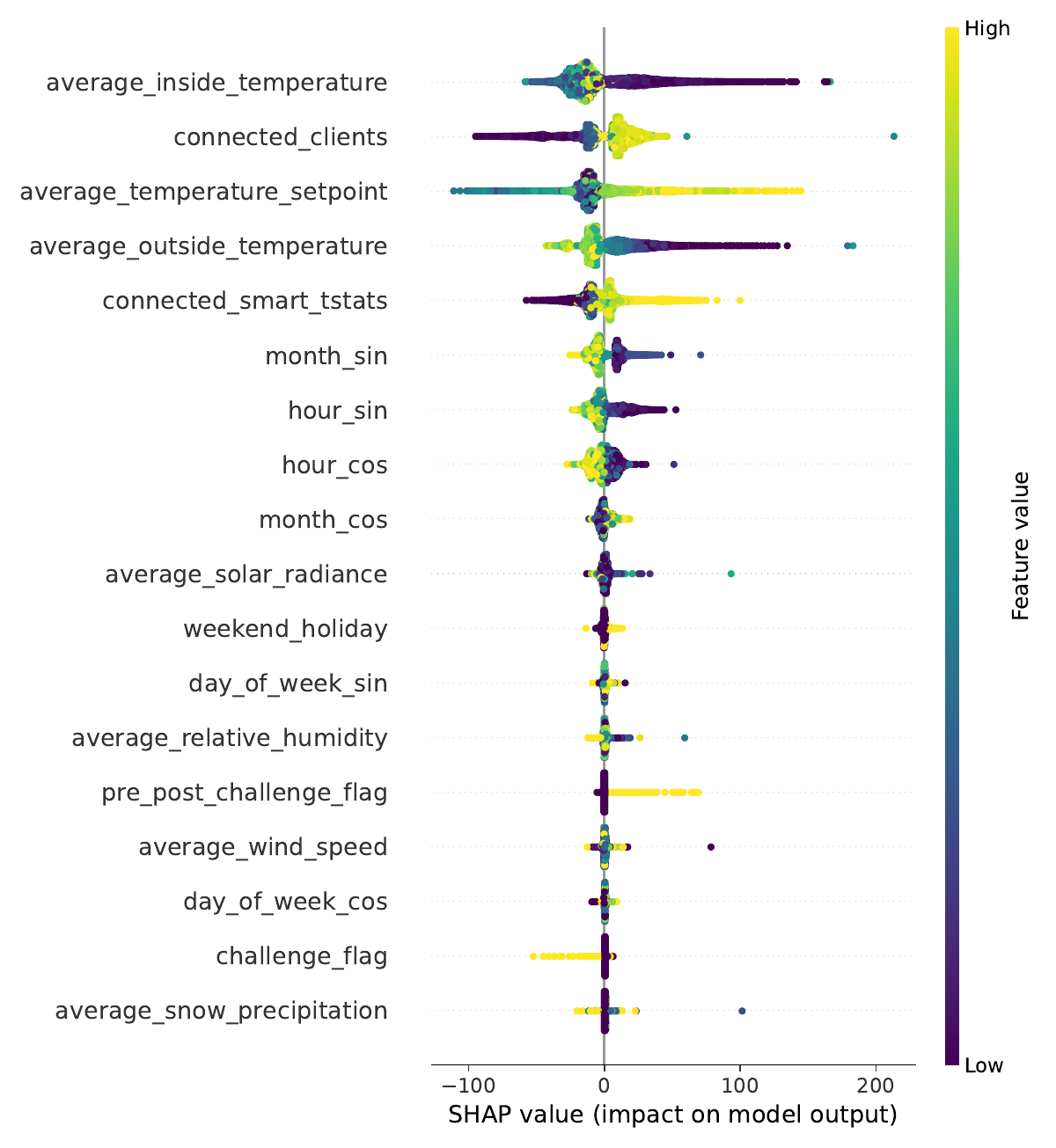}\vspace{-0.25cm}
         \caption{Substation B}\vspace{-0.25cm}
     \end{subfigure}
     \hfill
     
         \begin{subfigure}{1\linewidth}
         \centering
         \includegraphics[width=0.75\linewidth]{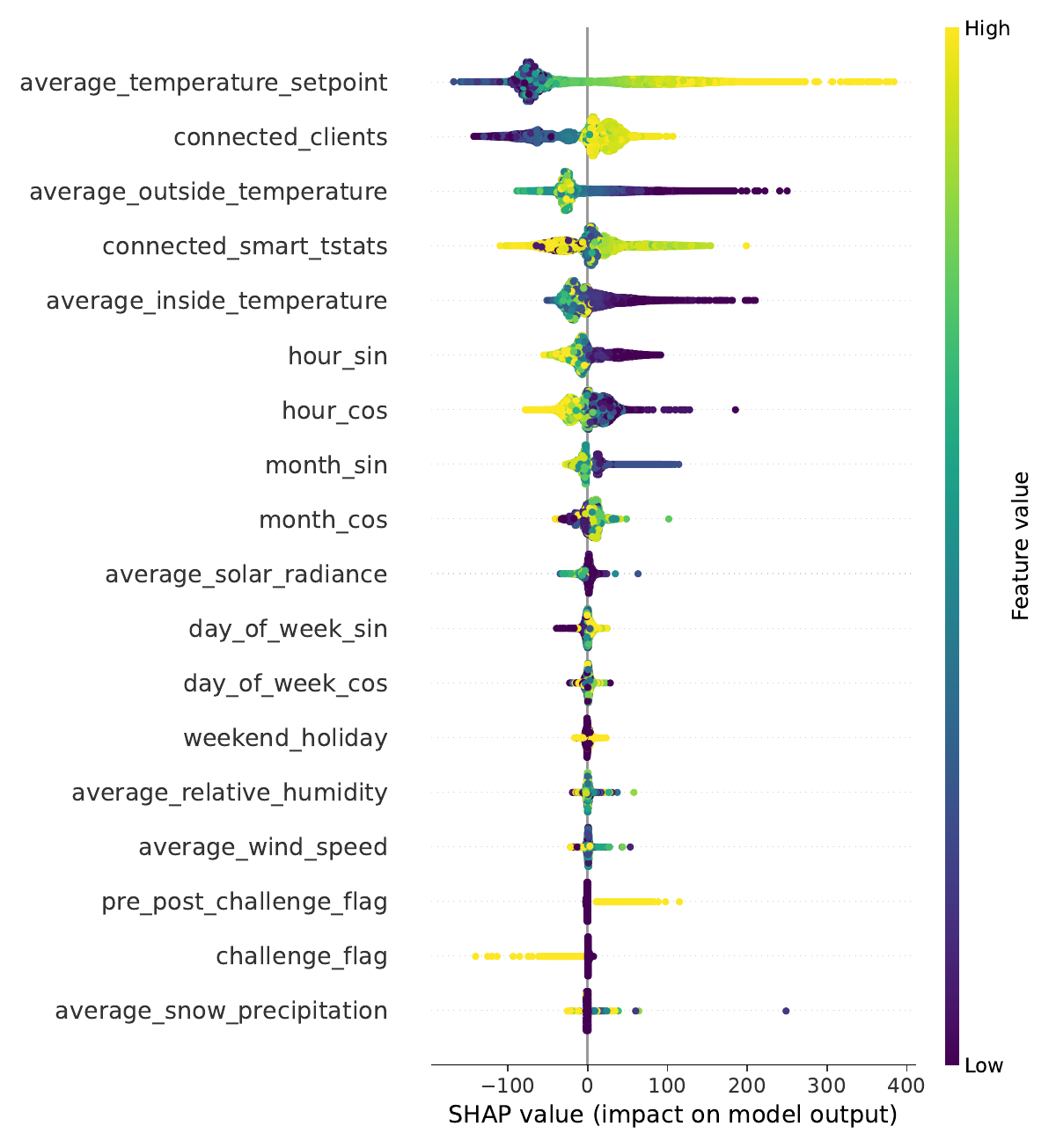}\vspace{-0.25cm}
         \caption{Substation C}
     \end{subfigure}
     \caption{Shapley analysis of key features for each substation on trained \texttt{XGBoost}}
     \label{fig:shap}
 \end{figure}

To have a more nuanced analysis of the contribution of each feature to the output, we also provide in Figure~\ref{fig:shap} an analysis of Shapley values of a trained extreme gradient boosting model (\texttt{XGBoost})~\cite{chen2016xgboost} for each substation. These analyses were realized with the Python package \texttt{SHAP}~\cite{lundberg2017SHAP}. In Figure~\ref{fig:shap}, we see that some lower-ranked features, viz., the challenge flags, sometimes have a strong impact on the model's output even though their general impact is null.

\begin{figure}[!htb]
     \centering
     \begin{subfigure}{1\linewidth}
         \includegraphics[width=\linewidth]{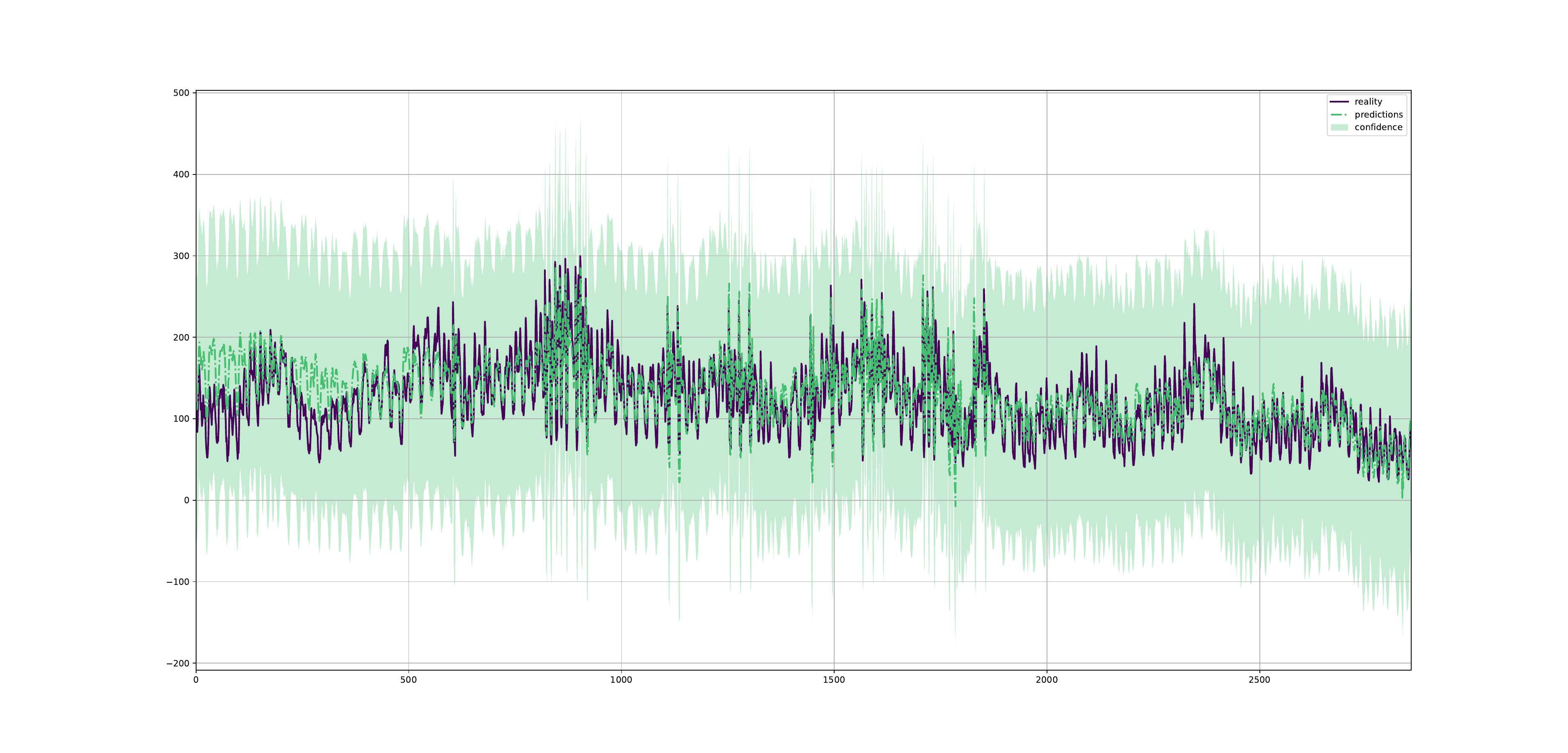}
         \caption{Substation A}
     \end{subfigure}
     \hfill
     \begin{subfigure}{1\linewidth}
         \includegraphics[width=\linewidth]{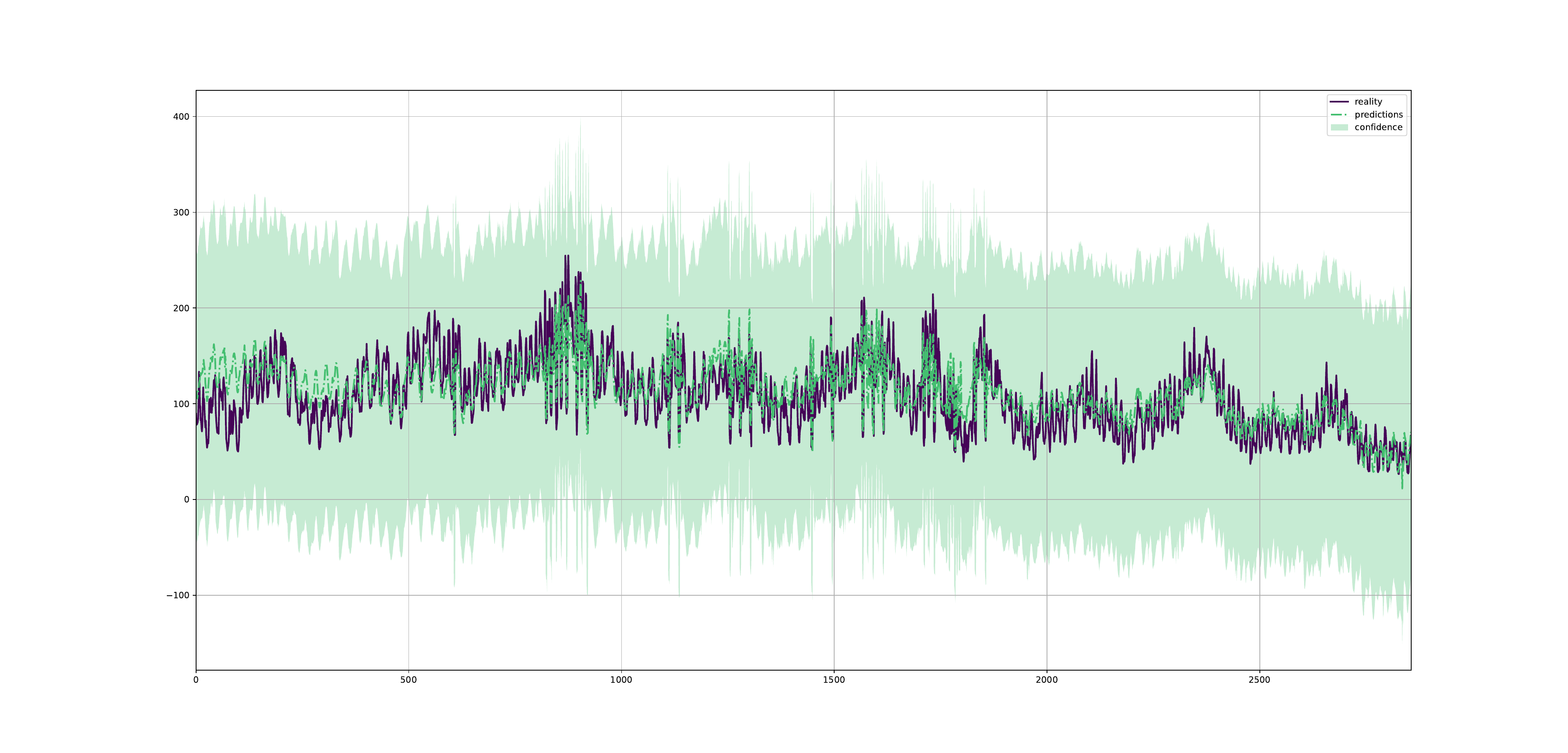}
         \caption{Substation B}
     \end{subfigure}
     \hfill
         \begin{subfigure}{1\linewidth}
         \includegraphics[width=\linewidth]{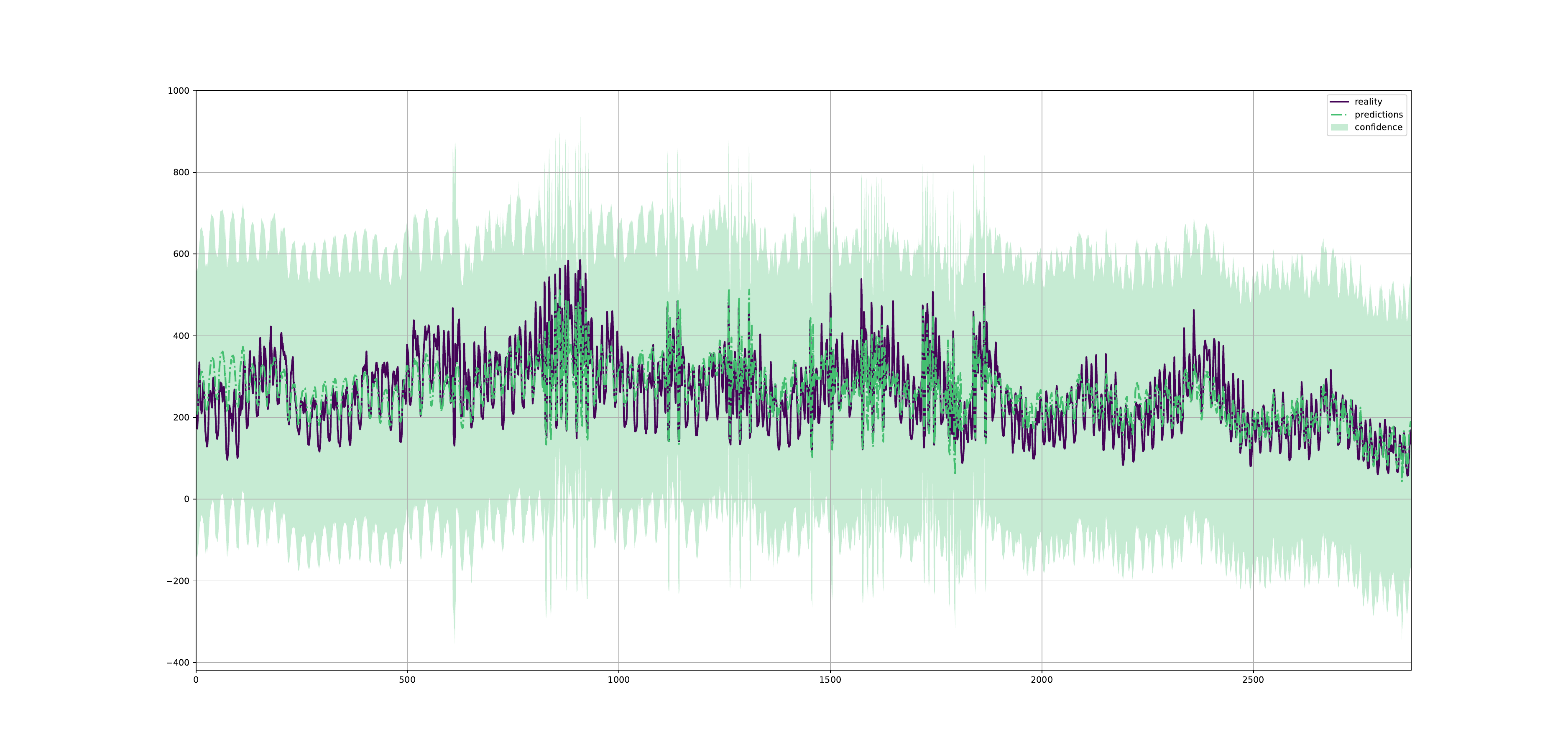}
         \caption{Substation C}
     \end{subfigure}
     \caption{Test predictions of the benchmark at each substation}
     \label{fig:preds_conso}
 \end{figure}

We conclude by presenting the test predictions for the best validation runs of each substation of the benchmark described in Section~\ref{sec:swfilter_experiments} in~Figure~\ref{fig:preds_conso}.

\end{document}